\pdfoutput=1

\documentclass[11pt]{article}

\usepackage[]{ACL2023}

\usepackage{times}
\usepackage{latexsym}
\usepackage{amsmath}
\usepackage{amssymb}

\usepackage{multirow} 
\usepackage[T1]{fontenc}

\usepackage[utf8]{inputenc}
\usepackage{algorithmic}
\usepackage{algorithm}
\usepackage{tcolorbox}\tcbuselibrary{skins}
\usepackage{microtype}

\usepackage{inconsolata}
\usepackage{graphicx}
\usepackage{amsmath}  
\usepackage{booktabs} 
\usepackage{microtype}
\usepackage{pgfplots}
\pgfplotsset{compat=1.18}
\usepgfplotslibrary{groupplots}
\usepackage{xcolor}
\usepackage{colortbl}

\usepackage{enumitem}

\newcommand{\AThreeS}{\textsc{A3S}\kern0.5pt}

\title{Can Activation Steering Capture Multidimensional Authorship Style?}

\author{
  Hieu Tran \quad Calvin Bao \quad Marine Carpuat \\
  University of Maryland, College Park \\
  \texttt{\{hieutt, csbao, marine\}@umd.edu}
}

\begin{document}
\maketitle
\begin{abstract}
Activation steering has shown promise for controlling LLM generation along well-defined attributes, but it remains unclear whether it can handle the multidimensional and hard-to-define nature of authorship style. We ask whether structured contrastive prompting along rhetorically-motivated dimensions can construct rich style representations directly in activation space, bypassing the need for natural language style descriptors or dedicated training. We find that the resulting directions share a common authorship backbone while conflicting on aspect-specific residuals that carry genuine stylistic signal, explaining why naive aggregation fails. We operationalize this in Aspect-Aware Activation Steering (\AThreeS), a training-free framework that merges per-aspect contrastive directions with interference-aware aggregation and tunes steering strength per instance. \AThreeS\ improves authorship style transfer where it is genuinely multi-aspect, outperforms a trained baseline in preference evaluations on out-of-domain benchmarks, and keeps target-exemplar overlap consistently low.

\end{abstract}

\newcommand{\ensuretext}[1]{#1}
\newcommand{\mcmarker}{\ensuretext{\textcolor{magenta}{\ensuremath{^{\textsc{M}}_{\textsc{C}}}}}}
\newcommand{\xnmarker}{\ensuretext{\textcolor{blue}{\ensuremath{^{\textsc{X}}_{\textsc{N}}}}}}
\newcommand{\wxmarker}{\ensuretext{\textcolor{cyan}{\ensuremath{^{\textsc{W}}_{\textsc{X}}}}}}
\newcommand{\mjmarker}{\ensuretext{\textcolor{cyan}{\ensuremath{^{\textsc{M}}_{\textsc{M}}}}}}
\newcommand{\kdmarker}{\ensuretext{\textcolor{purple}{\ensuremath{^{\textsc{K}}_{\textsc{D}}}}}}
\newcommand{\jpmmarker}{\ensuretext{\textcolor{green}{\ensuremath{^{\textsc{P}}_{\textsc{M}}}}}}
\newcommand{\aqmarker}{\ensuretext{\textcolor{green}{\ensuremath{^{\textsc{A}}_{\textsc{R}}}}}}
\newcommand{\samarker}{\ensuretext{\textcolor{purple}{\ensuremath{^{\textsc{S}}_{\textsc{A}}}}}}
\newcommand{\ebmarker}{\ensuretext{\textcolor{red}{\ensuremath{^{\textsc{E}}_{\textsc{B}}}}}}

\newcommand{\cbmarker}{\ensuretext{\textcolor{orange}{\ensuremath{^{\textsc{C}}_{\textsc{B}}}}}}

\newcommand{\cobmarker}{\ensuretext{\textcolor{violet}{\ensuremath{^{\textsc{Co}}_{\textsc{B}}}}}}

\newcommand{\remarker}{\ensuretext{\textcolor{orange}{\ensuremath{^{\textsc{R}}_{\textsc{e}}}}}}
\newcommand{\mycomment}[3]{\ensuretext{\textcolor{#3}{[#1 #2]}}}
\newcommand{\mc}[1]{\mycomment{\mcmarker}{#1}{magenta}}
\newcommand{\eb}[1]{\mycomment{\ebmarker}{#1}{red}}
\newcommand{\wx}[1]{\mycomment{\wxmarker}{#1}{cyan}}
\newcommand{\mjm}[1]{\mycomment{\mjmarker}{#1}{cyan}}
\newcommand{\sa}[1]{\mycomment{\samarker}{#1}{purple}}
\newcommand{\re}[1]{\mycomment{\remarker}{#1}{orange}}
\newcommand{\cb}[1]{\mycomment{\cbmarker}{#1}{orange}}

\newcommand{\cob}[1]{\mycomment{\cobmarker}{#1}{violet}}

\newcommand{\ignore}[1]{}
\newcommand{\aq}[1]{\mycomment{\aqmarker}{#1}{green}}

\section{Introduction}
Authorship style is multidimensional \citep{biber2009register}: a writer's identity is expressed through the interplay of lexical choice, sentence rhythm, rhetorical organization, and perspective, none of which can be reduced to a single attribute like formality or sentiment. This makes authorship style transfer (AST) a challenging task: a model must rewrite a source text to match the style of a target exemplar (e.g., a user's past writing) without altering the core message.

Previous work has primarily addressed this task via prompting or dedicated trained models. Prompt-based methods extract natural language style descriptors from the target exemplar and use them to guide rewriting \citep{patel2024lowresourceauthorshipstyletransfer, yang2025steeringlargelanguagemodels}. These methods require no training but flatten the fine-grained patterns that define individual authorship, and direct exemplar prompting raises apparent style match by reusing wording from the target rather than transferring its rhetorical signature. Dedicated models achieve stronger style transfer by conditioning generation on learned style embeddings \citep{horvitz-etal-2024-tinystyler}, but require training data and may not generalize well to new styles or out-of-domain settings.

A growing line of work shows that generation can be steered by intervening directly in a model's hidden states using directions extracted from contrastive activation statistics \citep{subramani-etal-2022-extracting, rimsky-etal-2024-steering, turner2024steeringlanguagemodelsactivation, chen2025personavectorsmonitoringcontrolling}. Prior steering work \citep{pmlr-v235-liu24bx, zhang-etal-2025-personalized, zhao2025steerxdisentangledsteeringllm} is usually designed for predefined attributes or learns a single contrastive direction for an entire target style. This is a poor fit for the multidimensional nature of authorship: collapsing all stylistic variation into one vector creates interference, where signals from different stylistic aspects may cancel one another.

In this paper, we ask: can the same LLM rewriting capability that prompting-based methods use to extract natural language style descriptors be used instead to construct richer representations directly in LLM activation space? Rather than compressing style into text and losing information in the process, we prompt the model to rewrite the target exemplar along rhetorically-motivated dimensions (\textit{Tone}, \textit{Perspective}, \textit{Structure}, \textit{Figurative Language}, and others, drawn from rhetoric and composition theory~\citep{abrams2014glossary, corbett1999classical, williams2017style}), and use the resulting rewrites to extract contrastive activation directions. This bypasses the bottleneck of natural language style description while avoiding the need for dedicated training.

We formalize this as the following hypothesis: \textit{structured contrastive prompting along rhetorically-motivated dimensions produces distinct and non-redundant directions in LLM activation space that collectively capture fine-grained authorship style better than any single direction.} We operationalize this hypothesis in Aspect-Aware Activation Steering (\AThreeS), a training-free framework that constructs per-aspect contrastive directions from a target exemplar, merges them with Parameter Competition Balancing (PCB)~\citep{du2024parameter} adapted from weight space to inference-time activation space, and tunes steering strength per instance with Hybrid Adaptive Search.

We test this hypothesis on three multi-aspect authorship style transfer benchmarks (MUD, LaMP, LongLaMP) 
The representational analysis reveals that per-aspect activation directions are not merely distinct: they share a common authorship backbone while conflicting on aspect-specific residuals that carry genuine stylistic signal, explaining why naive aggregation fails and motivating interference-aware merging. The benchmark results confirm the predictions of this analysis: A3S yields the largest gains over prompting baselines where authorship is genuinely multi-aspect, and outperforms TinyStyler in human and LLM preference evaluations, particularly on out-of-domain benchmarks. Our contributions are:
\begin{itemize}[leftmargin=1em]
    \item A geometric finding that exemplar-grounded aspect directions decompose into a shared authorship backbone plus aspect-specific residuals that carry genuine stylistic signal, providing a principled justification for multi-aspect activation steering in exemplar-based style transfer.
    \item \AThreeS, a training-free inference-time framework that operationalizes this finding via PCB-based aspect aggregation and per-instance Hybrid Adaptive Search.
    \item Empirical results showing that activation-space style representations constructed from rhetorical rewriting prompts capture fine-grained authorship style more effectively than natural language descriptors and generalize better out of domain than a learned style transfer model.
\end{itemize}

\section{Related Work}
\begin{figure*}[t]
    \centering
    \includegraphics[width=\textwidth]{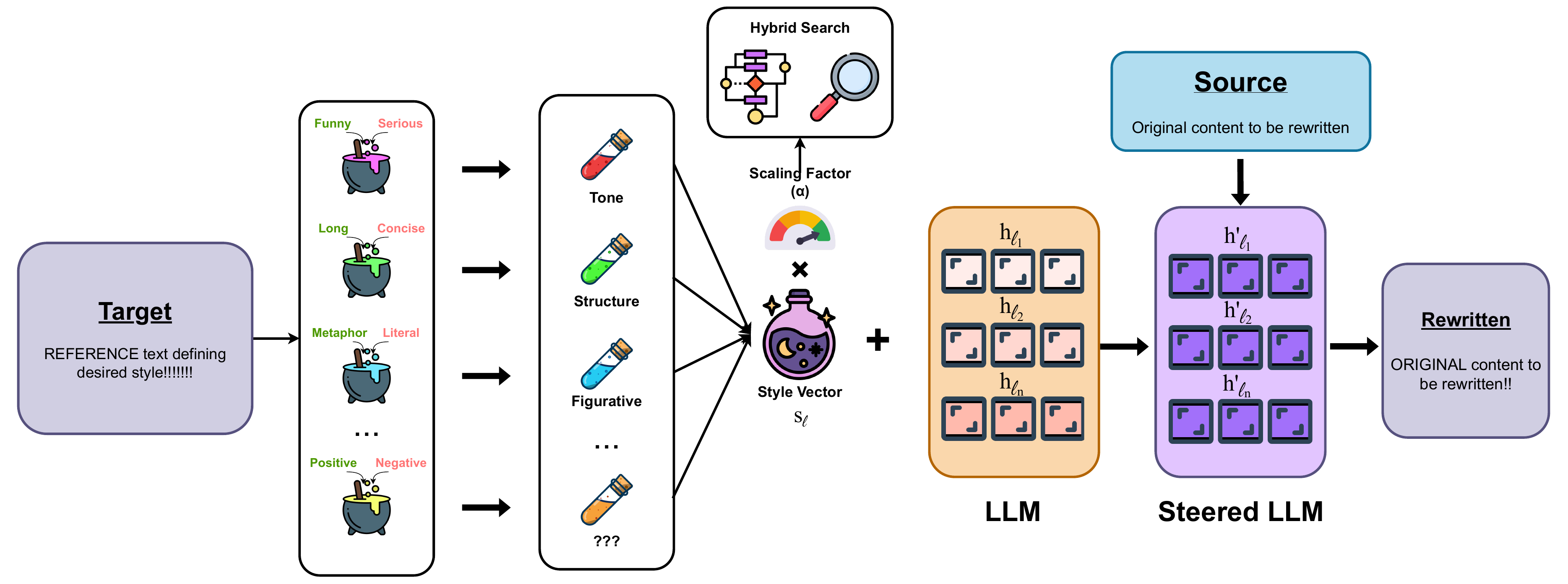}
    \caption{
    Overview of the proposed Aspect-Aware Activation Steering (\textsc{A3S}) framework. \textsc{A3S} decomposes the target exemplar into distinct rhetorical dimensions (e.g., Tone, Structure) via contrastive generation. These aspect vectors are synthesized using PCB-Merging to resolve interference, and injected with an intensity $\alpha^*$ dynamically tuned by Hybrid Adaptive Search.
    }
    \label{fig:a3s_architecture}
\end{figure*}

\textbf{Text Style Transfer} rewrites text to match a target style while preserving meaning~\citep{jin-etal-2022-deep}. Prior work largely focuses on attribute transfer over predefined properties such as sentiment or formality~\citep{rao-tetreault-2018-dear}, while recent work targets arbitrary or exemplar-based transfer, where style is defined implicitly by reference text. Prompt-based methods enable zero-shot transfer with instruction-tuned LLMs~\citep{reif-etal-2022-recipe}; STYLL~\citep{patel2024lowresourceauthorshipstyletransfer} and RG~\citep{yang2025steeringlargelanguagemodels} extract textual style descriptors to guide rewriting. These are easy to implement but offer limited style control, and direct exemplar prompting often inflates apparent style match by copying wordings rather than abstracting style. Training-based methods learn explicit transfer policies~\citep{liu-may-2025-style}, and TinyStyler~\cite{horvitz-etal-2024-tinystyler} conditions compact LMs on authorship embeddings for few-shot reconstruction. Our method also runs at inference time without parameter updates, but operates in activation space using per-aspect contrastive directions derived from the exemplar.

\textbf{Activation steering} modifies hidden states at inference time. Prior work shows global or contrastive vectors can steer sentiment, formality, or writing style without weight updates~\citep{konen-etal-2024-style, zhang-etal-2025-personalized, pmlr-v235-liu24bx}. These methods treat style as a single attribute with a fixed strength, a poor fit for user-specific authorship, which combines lexical, syntactic, and discourse-level signals no single direction expresses. Naive multi-aspect composition also fails due to directional interference~\citep{van-der-weij-etal-2024-extending}. \AThreeS~derives one direction per aspect, merges them with an interference-aware method, and picks a per-input steering coefficient.

\textbf{Authorship representation learning} produces dense embeddings of authorial style independent of topic, including LUAR~\citep{rivera-soto-etal-2021-learning}, style--content disentangled embeddings~\citep{wegmann-etal-2022-author}, and StyleDistance~\citep{patel-etal-2025-styledistance}. Such embeddings are used for attribution and for conditioning generation on a target author~\citep{horvitz-etal-2024-tinystyler}. We use them only for evaluation; steering instead operates over aspect-decomposed contrastive activations, which reveal geometric structure (\S\ref{sec:aspect_geometry}) not visible in a single dense embedding.

\textbf{Multi-attribute steering methods} address interference when steering multiple behaviors at once, typically over a predefined label set of alignment attributes (toxicity, bias, truthfulness) or coarse stylistic categories (tone). MAT-STEER~\citep{nguyen-etal-2025-multi} and MSRS~\citep{jiang2026adaptivemultisubspacerepresentationsteering} combine token-level gating with orthogonality on per-attribute subspaces; Dynamic Activation Composition~\citep{scalena-etal-2024-multi} and K-Steering~\citep{oozeer-etal-2025-beyond} adjust steering strength dynamically or train joint controllers. Our setting differs in two ways. First, the target is an arbitrary exemplar rather than a fixed label, so aspect directions are recomputed per style at inference time. Second, we hypothesize that structured contrastive prompting along rhetorically-motivated dimensions is a simple and effective way to construct rich style representations in activation space, and show that the resulting directions reveal a geometric structure that requires a targeted aspect-aggregation strategy.

\section{Methodology}

\subsection{Problem Formulation}
\label{sec:problem}

The central challenge in AST is not merely how to steer a model toward a target style, but how to represent that style from an exemplar without copying its content. We frame this as representation learning: rather than describing style in natural language, we construct activation-space directions that capture the exemplar's stylistic characteristics.

Given a source text $x$ and a target-author exemplar $t$ (a short reference or concatenation of recent posts), the goal is to generate $\hat{y}$ that preserves the meaning of $x$ while matching the style of $t$. We assume a frozen instruction-tuned LM $\mathcal{M}$ with no parameter updates and no supervised $\{(x_i, y_i)\}$ pairs in the target style; the only signal is the exemplar $t$. We treat the target as multi-aspect: rather than a single label, we model it as a tuple $(s_{\ell,1}, \dots, s_{\ell,|\mathcal{D}|})$ of activation directions over stylistic aspects $\mathcal{D}$, merged into a single steering vector at inference.

\subsection{Aspect-Aware Activation Steering (\AThreeS)}

We build on Contrastive Activation Steering (CAS), which identifies a direction vector separating two conditions in activation space. Let $h_\ell(y) \in \mathbb{R}^d$ denote the mean-pooled residual-stream representation of text $y$ at layer $\ell$. Given a positive text $y^+$ and a contrastive text $y^-$, CAS defines a steering direction as:
\begin{align}
s_\ell = h_\ell(y^+) - h_\ell(y^-).
\label{eq:cas}
\end{align}
During generation, hidden states are adjusted by the scaled direction:
\begin{align}
h'_\ell = h_\ell + \alpha s_\ell.
\label{eq:steering}
\end{align}
A larger $\alpha$ increases the stylistic effect but may reduce fluency or distort content.

\paragraph{Multi-aspect Style Vector Extraction.}
We decompose authorship style into dimensions $\mathcal{D}$ derived from literary criticism and composition \citep{abrams2014glossary, corbett1999classical, williams2017style}: \textit{Tone}, \textit{Diction}, \textit{Perspective}, \textit{Structure}, \textit{Figurative Language}, \textit{Syntax}, and \textit{Surface}.

\paragraph{Contrastive Pair Generation.}
For each dimension, we construct contrastive pairs $(y_d^+, y_d^-)$ grounded in the target exemplar. The positive text $y_d^+$ follows the exemplar style, while $y_d^-$ paraphrases the same content with the \emph{opposite} stylistic value (e.g., simple$\leftrightarrow$complex, figurative$\leftrightarrow$literal). We sample $K{=}4$ paraphrases per exemplar using prompts that ask the model to modify only one dimension while preserving meaning (Prompt templates in ~\ref{app:aspect_prompts}). Local aspects such as \textit{Tone} and \textit{Vocabulary} are reliably isolated by this procedure, while discourse aspects such as \textit{Structure} and \textit{Perspective} tend to produce noisier variants due to cross-aspect entanglement; we quantify this in \S\ref{sec:aspect_distinctness}.

For each dimension $d \in \mathcal{D}$ and sampled contrastive variant $k$, we compute:
\begin{align}
s_{\ell,d,k}
=
h_\ell(y_d^+) - h_\ell(y_{d,k}^-),
\end{align}
where $h_\ell(\cdot)$ is obtained by mean-pooling the residual-stream states over all tokens. We then aggregate the $K$ contrastive variants for each dimension:
\begin{align}
s_{\ell,d}
=
\frac{1}{K}\sum_{k=1}^{K}s_{\ell,d,k}.
\end{align}
Finally, we combine the aspect-specific directions by applying an aggregation function $f$:
\begin{align}
s_\ell = f\left( \{ s_{\ell, d} \}_{d \in \mathcal{D}} \right).
\end{align}
This extends one-dimensional activation steering \citep{zhang-etal-2025-personalized} by decomposing the target style into distinct stylistic aspects before merging their activation directions.

\paragraph{Style Aggregation Strategies.}
To synthesize aspect vectors into $s_\ell$, we use aggregation functions that can handle \textit{interference}, where aspects push in opposing directions. We compare:

\begin{enumerate}[leftmargin=1em]
    \item \textbf{Mean} and \textbf{Median}: element-wise aggregation over aspect vectors.
    
    \item \textbf{TIES-Merging~\cite{yadav2023tiesmerging}:} Trims low-magnitude values, elects a dominant sign, and merges only values aligned with that sign.
    
    \item \textbf{PCB-Merging~\cite{du2024parameter}:} Resolves conflicts via soft re-weighting using Intra-Balancing ($R_k$, relative magnitude) and Inter-Balancing ($A_k$, directional consistency). We compute $s_\ell = \sum_{k} (R_k + A_k) \odot s_k$, where:
    \begin{equation}
    \small
        R_k = \frac{|s_{\ell, k}|}{\sum_{j} |s_{\ell, j}| + \epsilon}, \quad
        A_k = \frac{s_{\ell, k} \cdot C_{\neq k}}{\|s_{\ell, k}\|_2 \|C_{\neq k}\|_2 + \epsilon}.
    \end{equation}
    This amplifies components that are both significant and consistent across aspects.
\end{enumerate}

As we will see in \S\ref{sec:aspect_distinctness}, exemplar-grounded aspect vectors $\{s_{\ell, d}\}$ decompose into a shared authorship backbone plus aspect-specific residuals that conflict pairwise yet carry genuine stylistic signal. This structure motivates PCB over naive averaging: mean averaging preserves the backbone but partially cancels the disagreeing residuals, whereas PCB re-weights components by both magnitude ($R_k$)  and inter-aspect consistency ($A_k$), preserving both components simultaneously.


\subsection{Adaptive Steering Coefficient Optimization}
\label{sec:adaptive_alpha}

Steering sensitivity varies across inputs and targets depending on the alignment of their activations with the style directions, so a fixed global $\alpha$ risks under-steering some inputs while pushing others into degeneration. We search for a sample-specific $\alpha^*$ to avoid this trade-off.

\paragraph{Degeneration Constraints (Hard Rejection).}
We define a validity function $\mathcal{V}(y_\alpha) \in \{0, 1\}$ and reject generations that violate:
\begin{enumerate}
    \item \textit{Suffix Periodicity:} a suffix of length $L$ repeats the preceding segment $y_{t-2L:t-L}$.
    \item \textit{Truncation:} the sequence reaches $T_{\max}$ without EOS and $P(\text{EOS}|y) < \epsilon$.
\end{enumerate}

\paragraph{Objective Function.}
Within the feasible region, we maximize steering intensity while penalizing distributional collapse. Let $P_b$ and $P_\alpha$ denote base and steered output distributions:
\begin{equation}
\label{eq:penalty}
\begin{aligned}
    \mathcal{P}_{\text{collapse}} = \, & \lambda_1 \text{CE}(P_b, P_\alpha) + \lambda_2 D_{\text{KL}}(P_b \parallel P_\alpha) \\
    & - \lambda_3 H(P_\alpha),
\end{aligned}
\end{equation}
where CE penalizes unlikely continuations, KL discourages excessive distributional drift, and entropy discourages mode collapse.

The steering objective $J(\alpha)$ is then defined as:
\begin{equation}
    J(\alpha) = \begin{cases} 
      -\infty & \text{if } \mathcal{V}(y_\alpha) = 0 \\
      \frac{\alpha \cdot |y_\alpha|}{1 + \mathcal{P}_{\text{collapse}}} & \text{otherwise}
   \end{cases}
\end{equation}
This favors stronger steering but discounts outputs whose logits diverge excessively from the base distribution.

\paragraph{Hybrid Adaptive Search.}
Since evaluating $J(\alpha)$ requires generation, we use a heuristic two-phase search:

\begin{enumerate}[leftmargin=1em]
    \item \textbf{Calibration:} a binary feasibility search over $[\alpha_{\min}, \alpha_{\max}]$ obtains a base coefficient $\alpha_{\text{base}}$.
    
    \item \textbf{Adaptation:} for each sample, we initialize at $\bar{\alpha} = \alpha_{\text{base}}$ and locally search for the largest valid value:
    \begin{itemize}
        \item \textit{Upward Probing:} If $\bar{\alpha}$ is valid, we iteratively step up ($\alpha + \delta$) until degeneration occurs, then backtrack.
        \item \textit{Downward Recovery:} If $\bar{\alpha}$ triggers hard rejection, we iteratively step down ($\alpha - \delta$) until validity is recovered.
    \end{itemize}
\end{enumerate}
Finally, we apply a brief golden-section search in the valid interval to refine $\alpha$.

\section{Experiments}


    

\paragraph{Datasets.}
We evaluate on three \emph{authorship style transfer} tasks spanning short-form and long-form generation: \textbf{Reddit Authorship (MUD)}, few-shot Reddit imitation following~\citep{patel2024lowresourceauthorshipstyletransfer, horvitz-etal-2024-tinystyler, yang2025steeringlargelanguagemodels}; \textbf{Twitter Authorship (LaMP)}, unsupervised transfer on Twitter user histories from the LaMP benchmark~\citep{salemi-etal-2024-lamp}; and \textbf{Long-Form Authorship (LongLaMP)}, transfer based on the Topic Writing task from LongLaMP~\citep{kumar2024longlampbenchmarkpersonalizedlongform}. Full construction details and sizes are in Appendix~\ref{app:datasets}.







\paragraph{Baselines.}
We compare \AThreeS\ against seven baselines spanning three methodological families.

\textit{Prompting baselines.}
We consider four prompting methods: Simple Prompting, which directly instructs the model to imitate the target style; STYLL~\citep{patel2024lowresourceauthorshipstyletransfer}, a multi-stage pipeline that generates style descriptors before rewriting; RG~\citep{yang2025steeringlargelanguagemodels}, which analyzes the target's linguistic register to guide rewriting; and Aspect-Aware Prompting, which elicits the same rhetorical aspects used by \AThreeS~in natural language (Appendix~\ref{app:aspect_aware_prompting}).

\textit{Activation-steering baselines.}
We evaluate two activation-based methods. \citet{konen-etal-2024-style} constructs layerwise directions by contrasting aggregated activations for the target style with those for other styles. Global Steering~\citep{zhang-etal-2025-personalized} constructs a single global direction from target-styled and neutral contrastive pairs.

\textit{Training-based baseline.}
TinyStyler~\citep{horvitz-etal-2024-tinystyler} is a few-shot method conditioning a pretrained LM on learned authorship embeddings. Because TinyStyler is trained on a subset of MUD, we treat MUD as its in-domain benchmark and LaMP, LongLaMP as out-of-domain benchmarks.

\paragraph{Evaluations.} We first automatically assess individual model outputs for:
\begin{itemize}[leftmargin=1em]
    \item \textbf{Style Matching.} We measure stylistic alignment using three embedding-based metrics: LUAR~\citep{rivera-soto-etal-2021-learning}, StyleCAV~\citep{wegmann-etal-2022-author}, and StyleDistance~\citep{patel-etal-2025-styledistance}. For each metric, we report \emph{Towards} (similarity to the target exemplar) and \emph{Away} (dissimilarity from the source) scores based on cosine similarity.
    \item \textbf{Meaning Preservation.} We measure semantic retention using MIS~\citep{babakov-etal-2022-large} and SBERT~\citep{reimers-gurevych-2019-sentence}.
    \item \textbf{Target Overlap.} To detect copying of the target exemplar, we report ROUGE-1/2/L scores between the generated output and the target text.
    \item \textbf{Length Shift.} To test whether gains are driven by verbosity rather than stylistic transfer, we report the character-length ratio $\mathrm{len}(y)/\mathrm{len}(x)$, the fraction of outputs more than 10\% longer than the source, and mean absolute length shift.
\end{itemize}

We also run a \textbf{pairwise preference evaluation} to compare the two best systems based on the above metrics, using a pairwise LLM-as-a-judge evaluation using GPT-4.1 on all benchmarks, and a human preference study on MUD. Full evaluation protocols are provided in Appendix~\ref{app:evaluation}.

\paragraph{Model Configuration.}
Main results use two base models: Llama-3.2-3B-Instruct~\cite{grattafiori2024llama3herdmodels} and Qwen3-4B~\cite{qwen3}; all analyses (\S\ref{sec:aspect_distinctness}--\S\ref{sec:ablation_alpha}) are conducted on Llama. Style vectors are injected at inference into a contiguous block of transformer layers: 8--18 for Llama and 11--31 for Qwen (layer analysis in Appendix~\ref{app:layer_analysis}). For all main results (Table~\ref{tab:main_results}), we use the top $K=4$ aspects (\textit{Figurative, Tone, Perspective, Structure}) selected via development-set ablations.

\section{Results and Analysis}

We evaluate A3S on three authorship style transfer benchmarks against seven baselines (\S\ref{sec:main_results}), then analyze the geometric structure of the aspect directions to explain why interference-aware aggregation is necessary (\S\ref{sec:aspect_distinctness}). Finally, \S\ref{sec:ablation_alpha} isolates the role of per-instance steering-strength search.

\subsection{Main Results}
\label{sec:main_results}

\begin{table*}[t]
\centering
\scriptsize
\renewcommand{\arraystretch}{0.92}
\setlength{\tabcolsep}{6pt}
\resizebox{\textwidth}{!}{%
\begin{tabular}{l*{6}{cc}}
\toprule
& \multicolumn{2}{c}{\textbf{LUAR}$\uparrow$}
& \multicolumn{2}{c}{\textbf{StyleCAV}$\uparrow$}
& \multicolumn{2}{c}{\textbf{StyleDistance}$\uparrow$}
& \multicolumn{2}{c}{\textbf{SBERT}$\uparrow$}
& \multicolumn{2}{c}{\textbf{ROUGE-L}$\downarrow$}
& \multicolumn{2}{c}{\textbf{Length}} \\
\cmidrule(lr){2-3}\cmidrule(lr){4-5}\cmidrule(lr){6-7}
\cmidrule(lr){8-9}\cmidrule(lr){10-11}\cmidrule(lr){12-13}
\textbf{Method}
& \textbf{Llama} & \textbf{Qwen}
& \textbf{Llama} & \textbf{Qwen}
& \textbf{Llama} & \textbf{Qwen}
& \textbf{Llama} & \textbf{Qwen}
& \textbf{Llama} & \textbf{Qwen}
& \textbf{Llama} & \textbf{Qwen} \\
\midrule

\rowcolor{gray!10}
\multicolumn{13}{l}{\textit{MUD (Reddit)}} \\
\textcolor{gray}{Simple}
& \textcolor{gray}{0.849} & \textcolor{gray}{0.770}
& \textcolor{gray}{0.833} & \textcolor{gray}{0.737}
& \textcolor{gray}{0.937} & \textcolor{gray}{0.894}
& \textcolor{gray}{0.293} & \textcolor{gray}{0.607}
& \textcolor{gray}{0.591} & \textcolor{gray}{0.411}
& \textcolor{gray}{23.45} & \textcolor{gray}{14.26} \\
STYLL
& 0.636 & 0.651 & 0.545 & 0.527 & 0.852 & 0.835
& 0.541 & 0.699 & 0.092 & 0.084 & 11.93 & 13.01 \\
RG
& 0.634 & 0.653 & 0.583 & 0.533 & 0.856 & 0.837
& 0.635 & \underline{0.796} & 0.080 & 0.074 & 10.24 & 25.56 \\
Aspect Prompt
& 0.654 & 0.653 & 0.558 & 0.478 & 0.850 & 0.826
& 0.597 & 0.662 & 0.111 & 0.113 & 13.70 & 17.30 \\
\cmidrule(lr){1-13}
Konen et al.
& 0.637 & 0.651 & 0.502 & 0.512 & 0.837 & 0.846
& 0.573 & 0.729 & 0.065 & 0.057 & 3.41 & 2.24 \\
Global Steering
& 0.626 & \underline{0.660} & 0.552 & 0.655 & 0.856 & 0.862
& 0.712 & 0.770 & 0.065 & 0.045 & 2.58 & 1.58 \\
TinyStyler$^{\dagger}$
& \multicolumn{2}{c}{0.648}
& \multicolumn{2}{c}{\textbf{0.788}}
& \multicolumn{2}{c}{\textbf{0.885}}
& \multicolumn{2}{c}{0.735}
& \multicolumn{2}{c}{0.068}
& \multicolumn{2}{c}{4.40} \\
\rowcolor{lightgray!25}
\textbf{\AThreeS}
& 0.656 & \textbf{0.665}
& 0.676 & \underline{0.678}
& 0.872 & \underline{0.873}
& 0.681 & \textbf{0.821}
& \textbf{0.039} & \underline{0.044}
& 0.88 & 1.26 \\
\midrule

\rowcolor{gray!10}
\multicolumn{13}{l}{\textit{LaMP (Tweet)}} \\
\textcolor{gray}{Simple}
& \textcolor{gray}{0.826} & \textcolor{gray}{0.814}
& \textcolor{gray}{0.821} & \textcolor{gray}{0.679}
& \textcolor{gray}{0.918} & \textcolor{gray}{0.866}
& \textcolor{gray}{0.472} & \textcolor{gray}{0.660}
& \textcolor{gray}{0.406} & \textcolor{gray}{0.428}
& \textcolor{gray}{4.40} & \textcolor{gray}{4.40} \\
STYLL
& 0.667 & 0.693 & 0.563 & 0.434 & 0.825 & 0.787
& 0.615 & 0.738 & 0.080 & 0.087 & 1.75 & 3.32 \\
RG
& 0.690 & 0.712 & 0.656 & 0.431 & 0.852 & 0.790
& 0.689 & \textbf{0.795} & 0.066 & 0.088 & 1.32 & 3.50 \\
Aspect Prompt
& 0.702 & \underline{0.741} & 0.634 & 0.415 & 0.844 & 0.778
& 0.678 & 0.708 & 0.094 & 0.110 & 2.57 & 3.11 \\
\cmidrule(lr){1-13}
Konen et al.
& 0.696 & 0.727 & 0.487 & 0.503 & 0.803 & 0.823
& 0.582 & 0.714 & 0.080 & 0.068 & 2.84 & 2.08 \\
Global Steering
& 0.685 & 0.740 & 0.567 & 0.737 & 0.833 & 0.881
& 0.704 & 0.707 & 0.080 & 0.052 & 2.08 & 1.20 \\
TinyStyler$^{\dagger}$
& \multicolumn{2}{c}{0.727}
& \multicolumn{2}{c}{\textbf{0.811}}
& \multicolumn{2}{c}{0.871}
& \multicolumn{2}{c}{0.746}
& \multicolumn{2}{c}{0.095}
& \multicolumn{2}{c}{2.33} \\
\rowcolor{lightgray!25}
\textbf{\AThreeS}
& 0.729 & \textbf{0.742}
& \underline{0.763} & 0.756
& \underline{0.886} & \textbf{0.896}
& 0.551 & \underline{0.748}
& \textbf{0.037} & \underline{0.050}
& 0.84 & 1.16 \\
\midrule

\rowcolor{gray!10}
\multicolumn{13}{l}{\textit{LongLaMP (Writing)}} \\
\textcolor{gray}{Simple}
& \textcolor{gray}{0.811} & \textcolor{gray}{0.767}
& \textcolor{gray}{0.773} & \textcolor{gray}{0.747}
& \textcolor{gray}{0.919} & \textcolor{gray}{0.902}
& \textcolor{gray}{0.304} & \textcolor{gray}{0.649}
& \textcolor{gray}{0.570} & \textcolor{gray}{0.370}
& \textcolor{gray}{1.65} & \textcolor{gray}{1.50} \\
STYLL
& 0.619 & 0.640 & 0.619 & 0.609 & 0.858 & 0.858
& 0.658 & 0.842 & 0.116 & 0.120 & 0.80 & 1.30 \\
RG
& 0.621 & 0.649 & 0.627 & 0.612 & 0.860 & 0.868
& 0.783 & \textbf{0.926} & 0.104 & 0.104 & 0.98 & 1.25 \\
Aspect Prompt
& 0.624 & 0.652 & 0.638 & 0.601 & 0.864 & 0.862
& 0.784 & \underline{0.916} & 0.109 & 0.103 & 1.00 & 1.56 \\
\cmidrule(lr){1-13}
Konen et al.
& 0.660 & \underline{0.665} & 0.646 & 0.648 & 0.865 & 0.869
& 0.654 & 0.813 & 0.123 & 0.114 & 1.06 & 0.94 \\
Global Steering
& 0.658 & 0.658 & 0.621 & 0.701 & 0.869 & \underline{0.877}
& 0.551 & 0.836 & \underline{0.088} & 0.103 & 0.92 & 0.97 \\
TinyStyler$^{\dagger}$
& \multicolumn{2}{c}{0.628}
& \multicolumn{2}{c}{\underline{0.714}}
& \multicolumn{2}{c}{0.875}
& \multicolumn{2}{c}{0.703}
& \multicolumn{2}{c}{\textbf{0.083}}
& \multicolumn{2}{c}{0.32} \\
\rowcolor{lightgray!25}
\textbf{\AThreeS}
& 0.663 & \textbf{0.674}
& 0.640 & \textbf{0.715}
& \underline{0.877} & \textbf{0.898}
& 0.788 & 0.863
& 0.114 & 0.103
& 0.78 & 1.00 \\
\bottomrule
\end{tabular}}
\caption{Results on three authorship-transfer benchmarks, with separate Llama and Qwen subcolumns. ROUGE-L measures target-exemplar overlap, and Length is the character-length ratio. \textbf{Bold}/\underline{underline} mark the best/second-best reported values per dataset; Simple is grayed because its high overlap indicates copying. $^{\dagger}$TinyStyler uses its own model and is conditioned on CAV embeddings, potentially favoring StyleCAV.}
\label{tab:main_results}
\end{table*}

\begin{figure}[t]
\centering
\includegraphics[width=0.9\columnwidth]{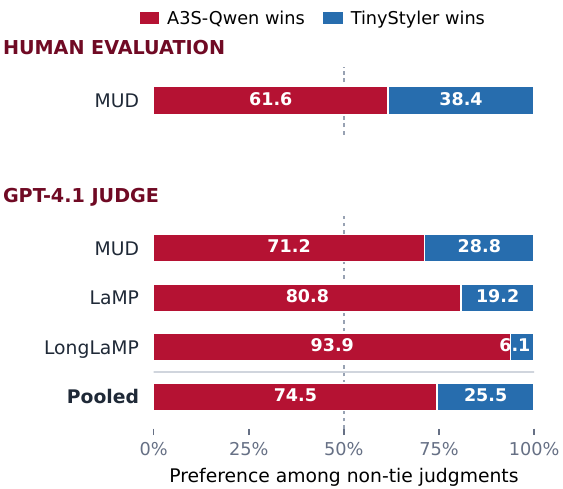}
\caption{Pairwise preference against TinyStyler. \AThreeS~(Qwen) is preferred on MUD by both human annotators and a GPT-4.1 judge. Preference is strongest on the out-of-domain LaMP and LongLaMP.}
\label{fig:winrate_combined}
\end{figure}

\paragraph{Similarity-based evaluation.}
Across tasks (Table~\ref{tab:main_results}), A3S leads on LUAR-based style fidelity, with \AThreeS\ (Qwen) obtaining the highest LUAR score on all three benchmarks. It also maintains near-source length ratios across both backbones and achieves the lowest target overlap on MUD and LaMP. Simple prompting achieves nominally high style scores but does so by copying the target exemplar, as evidenced by the high ROUGE-L scores; we gray those results out, as they reflect copying rather than style transfer. Structured prompting approaches (STYLL, RG) avoid this failure mode but fall short on style fidelity, suggesting that natural language style descriptors do not capture the fine-grained rhetorical patterns that define individual authorship.

Compared with the two activation-steering baselines, which steer style without aspect decomposition, \AThreeS\ improves all three style-fidelity metrics on MUD and LaMP across both backbones, demonstrating the value of aspect decomposition in activation space. On LongLaMP, \AThreeS\ (Qwen) also outperforms both baselines on all three style-fidelity metrics. These same-backbone comparisons demonstrate that the improvements cannot be attributed solely to using a stronger backbone.

A3S is competitive with TinyStyler on LUAR and StyleDistance across benchmarks, but trails on StyleCAV. Given that TinyStyler expands outputs by 2.3x to 4.4x on short-form tasks, and truncates by 0.3x on the long-form task, the style embedding gaps are difficult to interpret cleanly since longer outputs give embedding models more stylistic signal to match against. Moreover, TinyStyler uses embeddings trained with a CAV objective for conditioning and selecting its fine-tuning examples, which may favor its StyleCAV scores. Meaning preservation scores are mixed between the two systems.

The Qwen results strengthen this overall picture. \AThreeS~(Qwen) achieves the best LUAR scores across all three benchmarks and also leads on StyleDistance for LaMP and LongLaMP, while outperforming TinyStyler and both activation-steering baselines on SBERT in every setting ($0.821$ on MUD, $0.748$ on LaMP, and $0.863$ on LongLaMP). This suggests that the small SBERT drop observed for \AThreeS~(Llama) relative to TinyStyler is more likely a property of the underlying backbone than of the steering method itself. With a stronger base model, aspect-level steering is able to improve meaning preservation while still maintaining strong stylistic fidelity.

Together, these results demonstrate the benefit of decomposing style representations across rhetorical dimensions in activation space over both prompting and single-direction activation baselines. The comparison against TinyStyler is harder to resolve from automatic metrics alone due to length confounds, potential metric alignment, and mixed meaning preservation scores, which motivates a direct preference evaluation.\footnote{The appendices contain further details to contextualize the results. Qualitative examples in Table~\ref{tab:qualitative} illustrate the nature of the changes introduced by \AThreeS, while experiments on a control condition of single-attribute formality transfer (Appendix~\ref{app:gyafc}) confirm that A3S outperforms prompting baselines but offers diminishing returns over TinyStyler, consistent with the expectation that aspect decomposition adds the most value when target style is genuinely multidimensional.}



\paragraph{Human and LLM preference.}

To directly compare \AThreeS~(Qwen) and TinyStyler beyond automatic metrics, we ran a GPT-4.1 judgments over the full evaluation set (Figure~\ref{fig:winrate_combined}) and a human preference study on MUD, the domain TinyStyler was trained on, making this the toughest test for the unsupervised \AThreeS. On MUD, human annotators prefer \AThreeS\ in 114 of 185 non-tie judgments (61.6\% vs.\ 38.4\%), with 25 ties among 210 judgments. This preference is statistically significant ($p=0.0019$; 95\% item-level CI: $[53.3\%,69.7\%]$). The GPT-4.1 judge aligns with the human evaluation, also preferring \AThreeS\ on MUD and across all three benchmarks. Excluding ties, its win rates are 71.2\% on MUD ($p=2.7{\times}10^{-138}$), 80.8\% on LaMP ($p=2.7{\times}10^{-132}$), and 93.9\% on LongLaMP ($p=7.1{\times}10^{-21}$), with a pooled rate of 74.5\% (95\% Wilson CI: $[73.3\%,75.7\%]$). The consistent preference between human and LLM judgments supports that the improvements are perceptible to human readers and are not limited to automatic metrics, even on MUD where TinyStyler has an in-domain training advantage.

\subsection{Are Rhetorically-Motivated Dimensions Distinct in Activation Space?}
\label{sec:aspect_distinctness}
\label{sec:aspect_analysis}

\paragraph{Aspect purity.}
To check whether the LLM-generated contrastive variants actually isolate the intended aspect, we use GPT-4.1 to assign $0$--$5$ change scores along six stylistic dimensions for $200$ variants per target aspect, and report \emph{Purity@1} (target aspect is highest-scored) and \emph{Purity@2} (target is among top two) in Table~\ref{tab:aspect_purity_main}. Local aspects (Vocabulary, Tone, Syntax) reach Purity@1 above $0.87$. Discourse aspects fall sharply: Perspective $0.64$, Figurative $0.51$, Structure $0.26$. A lexical swap or short syntactic rewrite can leave the rest of the text alone, but changing argument structure or perspective tends to alter tone, syntax, and word choice at the same time, so the discourse variants pick up more cross-aspect noise. Structure still carries useful signal, even though it has the lowest StyleCAV in Table~\ref{tab:aspect_analysis}, but is one of the four aspects that PCB-Merging combines. Despite this entanglement at the text level, we find that the resulting activation directions have a clear geometric structure that makes their combination productive.

\begin{table}[h]
\centering
\small
\setlength{\tabcolsep}{6pt}
\begin{tabular}{lcc}
\toprule
\textbf{Target aspect} & \textbf{Purity@1} & \textbf{Purity@2} \\
\midrule
Vocabulary   & \textbf{0.985} & \textbf{1.000} \\
Tone         & 0.935 & 0.965 \\
Syntax       & 0.870 & 0.940 \\
Perspective  & 0.635 & 0.680 \\
Figurative   & 0.505 & 0.590 \\
Structure    & 0.255 & 0.510 \\
\bottomrule
\end{tabular}
\caption{Aspect isolation in LLM-generated rewrites.}
\label{tab:aspect_purity_main}
\end{table}

\begin{figure}[t]
\centering
\begin{minipage}{0.49\columnwidth}
\centering
\includegraphics[width=\linewidth]{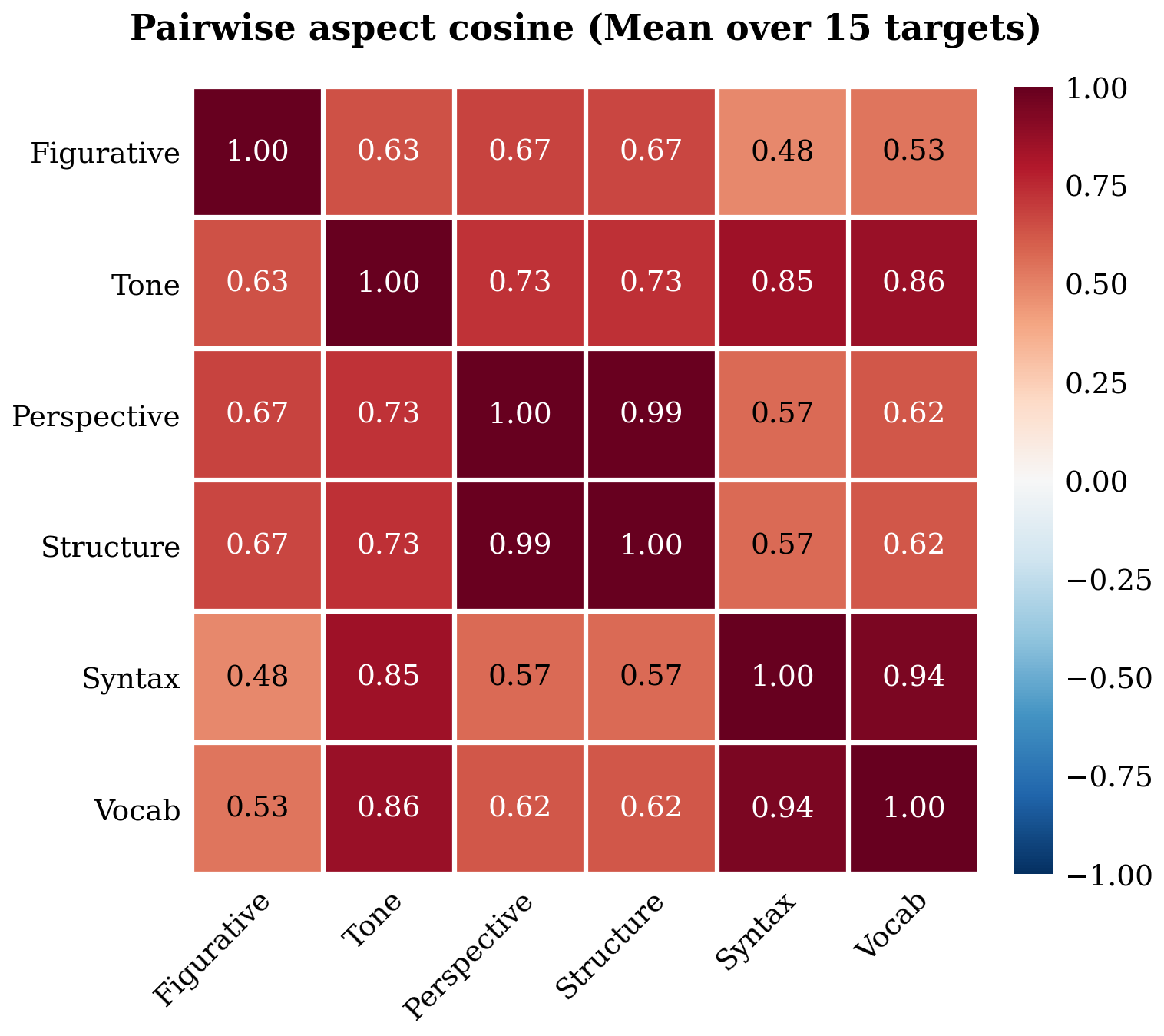}
\end{minipage}
\hfill
\begin{minipage}{0.49\columnwidth}
\centering
\includegraphics[width=\linewidth]{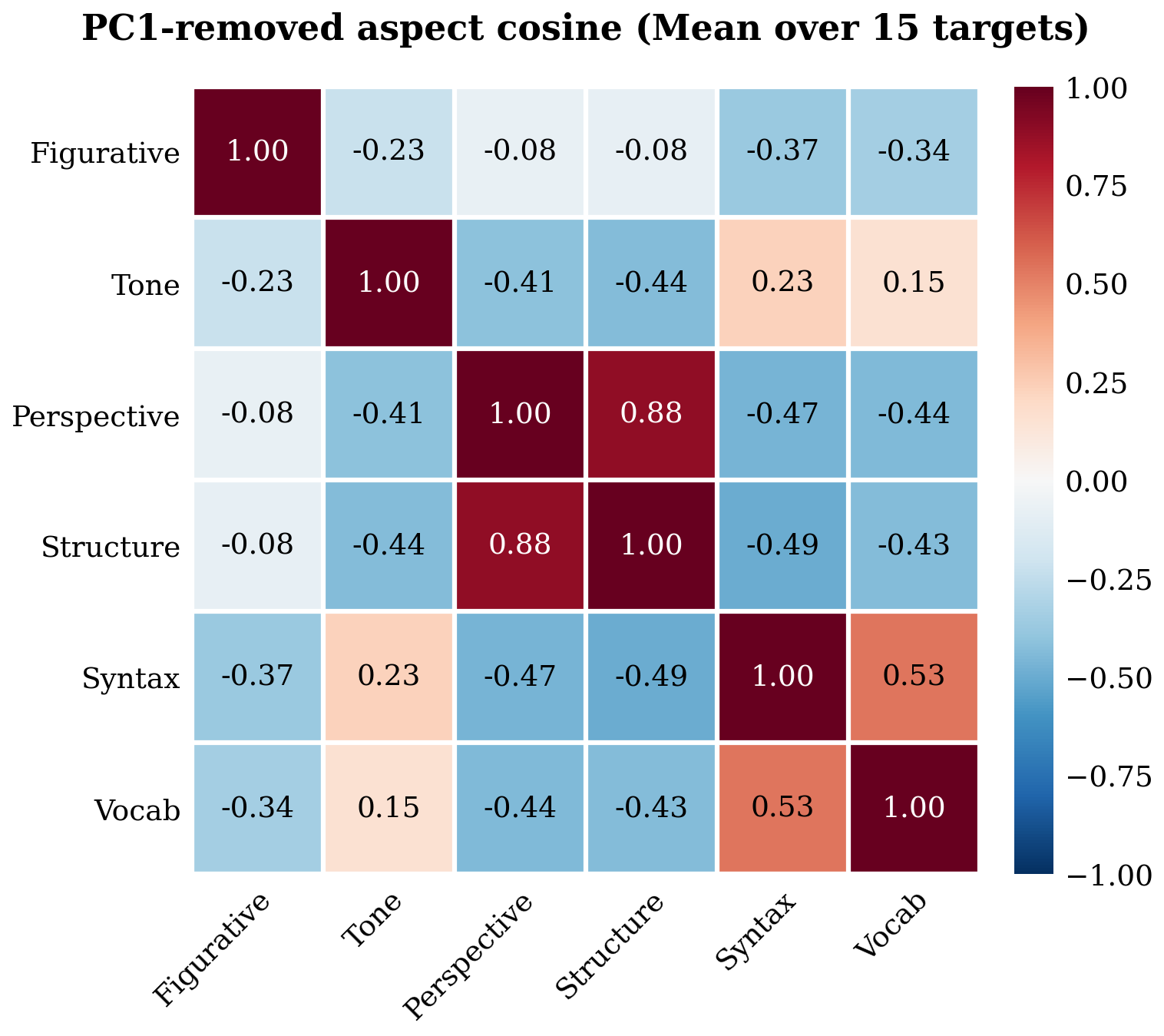}
\end{minipage}
\caption{Pairwise cosine between aspect directions (avg.\ 15 profiles). Raw aspects share a strong common component (left); after PC1 removal, residuals are negatively correlated in $60\%$ of pairs (right).}
\label{fig:aspect_cosine_main}
\end{figure}

\paragraph{Aspect geometry.}
\label{sec:aspect_geometry}
Figure~\ref{fig:aspect_cosine_main} shows the pairwise cosine between aspect directions averaged over 15 target profiles. In raw activation space (left), every pair is strongly positively aligned: the aspect vectors share a common component. After removing the first principal component (right), $60\%$ of pairs become negatively correlated, and the discourse aspects (\textit{Perspective}, \textit{Structure}) point opposite to the local aspects (\textit{Syntax}, \textit{Vocabulary}). The residuals carry $63\%$ of the per-aspect vector norm on average, so the disagreement is not a small perturbation around a shared direction.


\paragraph{Do aspect-specific residuals carry style distinction?} We test this in three ways. First, we sweep $\lambda$ in $u_{\mathrm{PC1}} + \lambda \cdot \mathrm{PCB}(R)$. Here, $u_{\mathrm{PC1}}$ is the unit vector along the first principal component of the aspect-specific activation directions and represents their shared authorship backbone, while $\lambda$ controls the strength and sign of the residual contribution. Thus, $\lambda{=}0$ gives PC1-only steering, $\lambda{<}0$ reverses the residual direction, and $\lambda{>}1$ amplifies it. Holding $u_{\mathrm{PC1}}$ fixed, StyleCAV generally increases from $0.153$ at $\lambda{=}-1$ to $0.359$ at $\lambda{=}1.5$, compared to $0.267$ for PC1-only steering (Table~\ref{tab:residual_analysis}, left). In other words, flipping the residual hurts and amplifying it helps, which confirms that the residuals are indeed aligned with the target style.

Second, the decomposition analysis in Table~\ref{tab:residual_analysis} (right) shows that residuals alone produce negative StyleCAV ($-0.218$ at $\alpha=0.4$), confirming they cannot function as standalone style directions and require the shared backbone as an anchor. However, combining PC1 with PCB-merged residuals outperforms combining PC1 with mean-merged residuals ($0.339$ vs. $0.315$ at $\alpha=0.5$), confirming that the residuals carry aspect-specific signal that mean averaging partially cancels.

\begin{table}[t]
\centering
\renewcommand{\arraystretch}{1.15}
\resizebox{\columnwidth}{!}{%
\begin{tabular}{lc @{\hspace{5mm}} lcc}
\cmidrule(r){1-2} \cmidrule(l){3-5}
$\lambda$ & \textbf{StyleCAV} & \textbf{Construction} & $\alpha{=}0.4$ & $\alpha{=}0.5$ \\
\cmidrule(r){1-2} \cmidrule(l){3-5}
$-1.00$ & 0.153 & PC1-only            & $\phantom{-}0.088$ & $\phantom{-}0.267$ \\
$-0.50$ & 0.192 & Resid.\ PCB         & $-0.218$           & $-0.166$           \\
\addlinespace
$\phantom{-}0.00$\,(PC1) & 0.267 & PC1{+}res.\ mean    & $\phantom{-}0.211$ & $\phantom{-}0.315$ \\
$\phantom{-}0.25$        & 0.294 & PC1{+}res.\ PCB     & $\phantom{-}0.241$ & $\phantom{-}0.339$ \\
\addlinespace
$\phantom{-}0.50$        & 0.310 & Raw mean            & $\phantom{-}0.211$ & $\phantom{-}0.317$ \\
$\phantom{-}0.75$        & 0.298 & \textbf{Raw PCB}    & \textbf{$\phantom{-}$0.287} & \textbf{$\phantom{-}$0.378} \\
\cmidrule(l){3-5}
$\phantom{-}1.00$        & 0.339 & & & \\
$\phantom{-}1.50$        & \textbf{0.359} & & & \\
\cmidrule(r){1-2}
\end{tabular}%
}
\vspace{1mm}
\caption{Residual role on MUD dev (StyleCAV $\uparrow$). \textit{Left}: $\lambda$-sweep of $v(\lambda){=}u_{\mathrm{PC1}}{+}\lambda\,\mathrm{PCB}(R)$. \textit{Right}: steering-vector constructions at fixed $\alpha$. Full results in Appendix~\ref{app:aspect_geometry}.}
\label{tab:residual_analysis}
\end{table}

Third, the aggregation ablation  shows that preserving the residuals algorithmically with PCB matters (Table~\ref{tab:aspect_analysis}). 
and explicit PC1 plus residual decomposition. 

\paragraph{Summary} Together, these results confirm that aspect-specific residuals carry genuine stylistic signal rather than noise, providing a mechanistic explanation for two patterns in Table~\ref{tab:main_results_full} and Figure~\ref{fig:winrate_combined}: (1) \AThreeS's consistent advantage over Global Steering, which collapses authorship style into a single direction and discards residual information entirely; and (2) the preference evaluation advantage over TinyStyler, particularly on out-of-domain benchmarks, where dense style embeddings generalize less well than the activation structure that \AThreeS constructs for each target style examplar.

\begin{table}[h]
\centering
\small
\setlength{\tabcolsep}{4pt}
\begin{tabular}{lcc}
\toprule
\textbf{Configuration} & \textbf{StyleCAV} ($\uparrow$) & \textbf{SBERT} ($\uparrow$) \\
\midrule
\multicolumn{3}{l}{\textit{Aggregation Strategies (Top 4 Aspects)}} \\
PCB-Merging & \textbf{0.485} & 0.694 \\
Median & 0.459 & \textbf{0.713} \\
Mean Averaging & 0.442 & 0.659 \\
TIES-Merging & 0.428 & 0.650 \\
\midrule
\multicolumn{3}{l}{\textit{Individual Dimensions}} \\
Figurative & 0.463 & 0.685 \\
Tone & 0.427 & 0.665 \\
Perspective & 0.403 & 0.713 \\
Structure & 0.399 & 0.750 \\
Syntax & 0.333 & 0.727 \\
Vocabulary & 0.310 & 0.724 \\
Surface & 0.103 & 0.692 \\
\midrule
\textit{Baseline} & & \\
Global Steering & 0.055 & 0.799 \\
\bottomrule
\end{tabular}
\caption{Aggregation strategies vs.\ single dimensions on MUD dev. PCB-Merging combines \textit{Figurative, Tone, Perspective, Structure}; Global is a single contrastive direction without aspect prompting.}
\label{tab:aspect_analysis}
\end{table}

\begin{figure}[t]
\centering
\includegraphics[width=0.9\columnwidth]{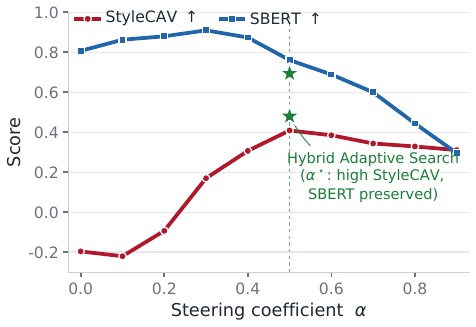}
\caption{Style--meaning trade-off on MUD dev as $\alpha$ varies. Fixed $\alpha$ must trade-off between under-steering (low StyleCAV) and degeneration (collapsing SBERT). Hybrid Adaptive Search ($\alpha^{\star}$, green stars) achieves StyleCAV $0.48$ while maintaining SBERT at $0.69$, outperforming all fixed $\alpha$ settings on the Pareto frontier.}
\label{fig:alpha_sweep}
\end{figure}

\subsection{Impact of Per-Instance Steering Strength}
\label{sec:ablation_alpha}

We compare Hybrid Adaptive Search against fixed $\alpha$ on MUD dev with the merged style vector (PCB over \textit{Figurative, Tone, Structure, Perspective}; Figure~\ref{fig:alpha_sweep}). StyleCAV increases as $\alpha$ grows and peaks near $\alpha{=}0.5$, after which SBERT drops sharply, so no single fixed $\alpha$ does well on both. Hybrid Adaptive Search matches the StyleCAV peak ($0.48$) while keeping SBERT at $0.69$, and is roughly $25\%$ faster than Binary Search. Construction overhead is modest when each target profile requires $16$ contrastive generations ($210$k--$478$k tokens, around $28$--$33$s; Appendix~\ref{app:quality_compute}). This cost is incurred once per author and then amortized across downstream queries.

\section{Conclusion}
We asked whether structured contrastive prompting along rhetorically-motivated dimensions can construct rich style representations in LLM activation space, sufficient for authorship style transfer without dedicated training. The answer is yes: the resulting directions are not just distinct but decompose into a shared authorship backbone plus aspect-specific residuals carrying genuine stylistic signal. This structure explains why naive aggregation fails and why interference-aware merging is necessary, and predicts the empirical pattern: largest gains where style combines multiple rhetorical dimensions, narrowing to parity with prompting or trained baselines when style reduces to a single attribute.

These findings suggest activation space contains rich structure for representing authorship style, accessible through structured contrastive prompting without parameter updates. Future work could ask whether the aspects can be discovered rather than imposed from rhetorical theory, and whether the geometric structure generalizes across model families and languages beyond English.


\section{Limitations}
\AThreeS~derives its aspect directions from contrastive variants generated by instruction-tuned LLMs. This introduces two important assumptions into our analysis. First, the quality of the aspect directions depends on the quality of the generated variants. If the generator cannot vary one aspect while keeping the others relatively fixed, the resulting aspect direction will also reflect changes in other aspects. We quantify this effect for discourse-level aspects in \S\ref{sec:aspect_analysis}. Second, the shared PC1 backbone we observe could partly reflect the properties of the variant generator itself rather than authorship alone. As a result, the geometric findings should be interpreted as joint properties of the steered model and the generator used to construct the variants. Replicating the analysis with alternative generators would, therefore, be an important direction for future work.

The datasets are primarily English, so cross-lingual transfer of the aspect geometry is untested. In addition, \AThreeS~introduces nontrivial inference-time overhead. Direction construction requires $K{=}16$ generations per target profile, and the per-instance $\alpha$ search adds several additional forward passes for each input. Both costs could likely be reduced by steering with fewer aspects, although this would come at some loss in quality.


\section{Acknowledgments}
This research is supported in part by the Office of the Director of National Intelligence (ODNI), Intelligence Advanced Research Projects Activity (IARPA), via the HIATUS Program contract \#2022-22072200006. The views and conclusions contained herein are those of the authors and should not be interpreted as necessarily representing the official policies, either expressed or implied, of ODNI, IARPA, or the U.S. Government. The U.S. Government is authorized to reproduce and distribute reprints for governmental purposes notwithstanding any copyright annotation therein.

\subsection*{AI Use Statement}
Generative AI tools were used to assist with drafting and editing the paper.
The authors reviewed the resulting text, equations, citations, and code-derived
descriptions and take responsibility for the final content.

\bibliography{anthology,references}

@misc{yang2025steeringlargelanguagemodels,
      title={Steering Large Language Models with Register Analysis for Arbitrary Style Transfer}, 
      author={Xinchen Yang and Marine Carpuat},
      year={2025},
      eprint={2505.00679},
      archivePrefix={arXiv},
      primaryClass={cs.CL},
      url={https://arxiv.org/abs/2505.00679}, 
}

@misc{patel2024lowresourceauthorshipstyletransfer,
      title={Low-Resource Authorship Style Transfer: Can Non-Famous Authors Be Imitated?}, 
      author={Ajay Patel and Nicholas Andrews and Chris Callison-Burch},
      year={2024},
      eprint={2212.08986},
      archivePrefix={arXiv},
      primaryClass={cs.CL},
      url={https://arxiv.org/abs/2212.08986}, 
}

@inproceedings{zhang-etal-2025-personalized,
    title = "Personalized Text Generation with Contrastive Activation Steering",
    author = "Zhang, Jinghao  and
      Liu, Yuting  and
      Wang, Wenjie  and
      Liu, Qiang  and
      Wu, Shu  and
      Wang, Liang  and
      Chua, Tat-Seng",
    editor = "Che, Wanxiang  and
      Nabende, Joyce  and
      Shutova, Ekaterina  and
      Pilehvar, Mohammad Taher",
    booktitle = "Proceedings of the 63rd Annual Meeting of the Association for Computational Linguistics (Volume 1: Long Papers)",
    month = jul,
    year = "2025",
    address = "Vienna, Austria",
    publisher = "Association for Computational Linguistics",
    url = "https://aclanthology.org/2025.acl-long.353/",
    doi = "10.18653/v1/2025.acl-long.353",
    pages = "7128--7141",
    ISBN = "979-8-89176-251-0"
}

@misc{zhao2025steerxdisentangledsteeringllm,
      title={SteerX: Disentangled Steering for LLM Personalization}, 
      author={Xiaoyan Zhao and Ming Yan and Yilun Qiu and Haoting Ni and Yang Zhang and Fuli Feng and Hong Cheng and Tat-Seng Chua},
      year={2025},
      eprint={2510.22256},
      archivePrefix={arXiv},
      primaryClass={cs.CL},
      url={https://arxiv.org/abs/2510.22256}, 
}

@misc{turner2024steeringlanguagemodelsactivation,
      title={Steering Language Models With Activation Engineering}, 
      author={Alexander Matt Turner and Lisa Thiergart and Gavin Leech and David Udell and Juan J. Vazquez and Ulisse Mini and Monte MacDiarmid},
      year={2024},
      eprint={2308.10248},
      archivePrefix={arXiv},
      primaryClass={cs.CL},
      url={https://arxiv.org/abs/2308.10248}, 
}

@inproceedings{rimsky-etal-2024-steering,
    title = "Steering Llama 2 via Contrastive Activation Addition",
    author = "Rimsky, Nina  and
      Gabrieli, Nick  and
      Schulz, Julian  and
      Tong, Meg  and
      Hubinger, Evan  and
      Turner, Alexander",
    editor = "Ku, Lun-Wei  and
      Martins, Andre  and
      Srikumar, Vivek",
    booktitle = "Proceedings of the 62nd Annual Meeting of the Association for Computational Linguistics (Volume 1: Long Papers)",
    month = aug,
    year = "2024",
    address = "Bangkok, Thailand",
    publisher = "Association for Computational Linguistics",
    url = "https://aclanthology.org/2024.acl-long.828/",
    doi = "10.18653/v1/2024.acl-long.828",
    pages = "15504--15522"
}

@misc{khan2021deepmetriclearningapproach,
      title={A Deep Metric Learning Approach to Account Linking}, 
      author={Aleem Khan and Elizabeth Fleming and Noah Schofield and Marcus Bishop and Nicholas Andrews},
      year={2021},
      eprint={2105.07263},
      archivePrefix={arXiv},
      primaryClass={cs.SI},
      url={https://arxiv.org/abs/2105.07263}, 
}

@misc{chen2025personavectorsmonitoringcontrolling,
      title={Persona Vectors: Monitoring and Controlling Character Traits in Language Models}, 
      author={Runjin Chen and Andy Arditi and Henry Sleight and Owain Evans and Jack Lindsey},
      year={2025},
      eprint={2507.21509},
      archivePrefix={arXiv},
      primaryClass={cs.CL},
      url={https://arxiv.org/abs/2507.21509}, 
}

@inproceedings{konen-etal-2024-style,
    title = "Style Vectors for Steering Generative Large Language Models",
    author = {Konen, Kai  and
      Jentzsch, Sophie  and
      Diallo, Diaoul{\'e}  and
      Sch{\"u}tt, Peer  and
      Bensch, Oliver  and
      El Baff, Roxanne  and
      Opitz, Dominik  and
      Hecking, Tobias},
    editor = "Graham, Yvette  and
      Purver, Matthew",
    booktitle = "Findings of the Association for Computational Linguistics: EACL 2024",
    month = mar,
    year = "2024",
    address = "St. Julian{'}s, Malta",
    publisher = "Association for Computational Linguistics",
    url = "https://aclanthology.org/2024.findings-eacl.52/",
    pages = "782--802"
}

@misc{grattafiori2024llama3herdmodels,
      title={The Llama 3 Herd of Models}, 
      author={Llama, Team},
      year={2024},
      eprint={2407.21783},
      archivePrefix={arXiv},
      primaryClass={cs.AI},
      url={https://arxiv.org/abs/2407.21783}, 
}

@InProceedings{pmlr-v235-liu24bx,
  title     = {In-context Vectors: Making In Context Learning More Effective and Controllable Through Latent Space Steering},
  author    = {Liu, Sheng and Ye, Haotian and Xing, Lei and Zou, James Y.},
  booktitle = {Proceedings of the 41st International Conference on Machine Learning},
  year      = {2024},
  pages     = {32287--32307},
  volume    = {235},
  series    = {Proceedings of Machine Learning Research},
  publisher = {PMLR},
  url       = {https://proceedings.mlr.press/v235/liu24bx.html}
}

@inproceedings{nguyen-etal-2025-multi,
  title     = {Multi-Attribute Steering of Language Models via Targeted Intervention},
  author    = {Nguyen, Duy and Prasad, Archiki and Stengel-Eskin, Elias and Bansal, Mohit},
  booktitle = {Proceedings of the 63rd Annual Meeting of the Association for Computational Linguistics (Volume 1: Long Papers)},
  year      = {2025},
  pages     = {20619--20634},
  doi       = {10.18653/v1/2025.acl-long.1007},
  url       = {https://aclanthology.org/2025.acl-long.1007/}
}

@inproceedings{scalena-etal-2024-multi,
  title     = {Multi-property Steering of Large Language Models with Dynamic Activation Composition},
  author    = {Scalena, Daniel and Sarti, Gabriele and Nissim, Malvina},
  booktitle = {Proceedings of the 7th BlackboxNLP Workshop: Analyzing and Interpreting Neural Networks for NLP},
  year      = {2024},
  pages     = {577--603},
  doi       = {10.18653/v1/2024.blackboxnlp-1.34},
  url       = {https://aclanthology.org/2024.blackboxnlp-1.34/}
}

@inproceedings{oozeer-etal-2025-beyond,
  title = "Beyond Linear Steering: Unified Multi-Attribute Control for Language Models",
  author = "Oozeer, Narmeen Fatimah and Marks, Luke and Barez, Fazl and Abdullah, Amir",
  booktitle = "Findings of the Association for Computational Linguistics: EMNLP 2025",
  month = nov,
  year = "2025",
  address = "Suzhou, China",
  publisher = "Association for Computational Linguistics",
  url = "https://aclanthology.org/2025.findings-emnlp.1278/",
  doi = "10.18653/v1/2025.findings-emnlp.1278",
  pages = "23513--23557"
}

@inproceedings{liu-may-2025-style,
  title = "Style Transfer with Multi-iteration Preference Optimization",
  author = "Liu, Shuai and May, Jonathan",
  booktitle = "Proceedings of the 2025 Conference of the Nations of the Americas Chapter of the Association for Computational Linguistics: Human Language Technologies (Volume 1: Long Papers)",
  month = apr,
  year = "2025",
  address = "Albuquerque, New Mexico",
  publisher = "Association for Computational Linguistics",
  url = "https://aclanthology.org/2025.naacl-long.135/",
  doi = "10.18653/v1/2025.naacl-long.135",
  pages = "2663--2681"
}

@misc{van-der-weij-etal-2024-extending,
  title = {Extending Activation Steering to Broad Skills and Multiple Behaviours},
  author = {Teun van der Weij and Massimo Poesio and Nandi Schoots},
  year = {2024},
  eprint = {2403.05767},
  archivePrefix = {arXiv},
  primaryClass = {cs.LG},
  url = {https://arxiv.org/abs/2403.05767}
}

@misc{jiang2026adaptivemultisubspacerepresentationsteering,
      title={Adaptive Multi-Subspace Representation Steering for Attribute Alignment in Large Language Models}, 
      author={Xinyan Jiang and Lin Zhang and Jiayi Zhang and Qingsong Yang and Guimin Hu and Di Wang and Lijie Hu},
      year={2026},
      eprint={2508.10599},
      archivePrefix={arXiv},
      primaryClass={cs.AI},
      url={https://arxiv.org/abs/2508.10599}, 
}

@inproceedings{
yadav2023tiesmerging,
title={{TIES}-Merging: Resolving Interference When Merging Models},
author={Prateek Yadav and Derek Tam and Leshem Choshen and Colin Raffel and Mohit Bansal},
booktitle={Thirty-seventh Conference on Neural Information Processing Systems},
year={2023},
url={https://openreview.net/forum?id=xtaX3WyCj1}
}

@inproceedings{
    du2024parameter,
    title={Parameter Competition Balancing for Model Merging},
    author={Guodong DU and Junlin Lee and Jing Li and Runhua Jiang and Yifei Guo and Shuyang Yu and Hanting Liu and Sim Kuan Goh and Ho-Kin Tang and Daojing He and Min Zhang},
    booktitle={The Thirty-eighth Annual Conference on Neural Information Processing Systems},
    year={2024},
    url={https://openreview.net/forum?id=l5SbrtvSRS}
}

@book{abrams2014glossary,
  title={A Glossary of Literary Terms},
  author={Abrams, M.H. and Harpham, G.},
  isbn={9781285974514},
  url={https://books.google.com/books?id=X7-iAgAAQBAJ},
  year={2014},
  publisher={Cengage Learning}
}

@book{corbett1999classical,
  title={Classical Rhetoric for the Modern Student},
  author={Corbett, E.P.J. and Connors, R.J.},
  isbn={9780195115420},
  lccn={97026686},
  url={https://books.google.com/books?id=JV_hwAEACAAJ},
  year={1999},
  publisher={Oxford University Press}
}

@book{williams2017style,
  title={Style: Lessons in Clarity and Grace},
  author={Williams, J.M. and Bizup, J.},
  isbn={9780134080413},
  lccn={2015031148},
  series={Always learning},
  url={https://books.google.com/books?id=XXpSjgEACAAJ},
  year={2017},
  publisher={Pearson}
}

@inproceedings{horvitz-etal-2024-tinystyler,
    title = "{T}iny{S}tyler: Efficient Few-Shot Text Style Transfer with Authorship Embeddings",
    author = "Horvitz, Zachary  and
      Patel, Ajay  and
      Singh, Kanishk  and
      Callison-Burch, Chris  and
      McKeown, Kathleen  and
      Yu, Zhou",
    editor = "Al-Onaizan, Yaser  and
      Bansal, Mohit  and
      Chen, Yun-Nung",
    booktitle = "Findings of the Association for Computational Linguistics: EMNLP 2024",
    month = nov,
    year = "2024",
    address = "Miami, Florida, USA",
    publisher = "Association for Computational Linguistics",
    url = "https://aclanthology.org/2024.findings-emnlp.781/",
    doi = "10.18653/v1/2024.findings-emnlp.781",
    pages = "13376--13390"
}

@misc{kumar2024longlampbenchmarkpersonalizedlongform,
      title={LongLaMP: A Benchmark for Personalized Long-form Text Generation}, 
      author={Ishita Kumar and Snigdha Viswanathan and Sushrita Yerra and Alireza Salemi and Ryan A. Rossi and Franck Dernoncourt and Hanieh Deilamsalehy and Xiang Chen and Ruiyi Zhang and Shubham Agarwal and Nedim Lipka and Chien Van Nguyen and Thien Huu Nguyen and Hamed Zamani},
      year={2024},
      eprint={2407.11016},
      archivePrefix={arXiv},
      primaryClass={cs.CL},
      url={https://arxiv.org/abs/2407.11016}, 
}

@inproceedings{salemi-etal-2024-lamp,
    title = "{L}a{MP}: When Large Language Models Meet Personalization",
    author = "Salemi, Alireza  and
      Mysore, Sheshera  and
      Bendersky, Michael  and
      Zamani, Hamed",
    editor = "Ku, Lun-Wei  and
      Martins, Andre  and
      Srikumar, Vivek",
    booktitle = "Proceedings of the 62nd Annual Meeting of the Association for Computational Linguistics (Volume 1: Long Papers)",
    month = aug,
    year = "2024",
    address = "Bangkok, Thailand",
    publisher = "Association for Computational Linguistics",
    url = "https://aclanthology.org/2024.acl-long.399/",
    doi = "10.18653/v1/2024.acl-long.399",
    pages = "7370--7392"
}

@inproceedings{patel-etal-2025-styledistance,
    title = "{S}tyle{D}istance: Stronger Content-Independent Style Embeddings with Synthetic Parallel Examples",
    author = "Patel, Ajay  and
      Zhu, Jiacheng  and
      Qiu, Justin  and
      Horvitz, Zachary  and
      Apidianaki, Marianna  and
      McKeown, Kathleen  and
      Callison-Burch, Chris",
    editor = "Chiruzzo, Luis  and
      Ritter, Alan  and
      Wang, Lu",
    booktitle = "Proceedings of the 2025 Conference of the Nations of the Americas Chapter of the Association for Computational Linguistics: Human Language Technologies (Volume 1: Long Papers)",
    month = apr,
    year = "2025",
    address = "Albuquerque, New Mexico",
    publisher = "Association for Computational Linguistics",
    url = "https://aclanthology.org/2025.naacl-long.436/",
    doi = "10.18653/v1/2025.naacl-long.436",
    pages = "8662--8685",
    ISBN = "979-8-89176-189-6"
}

@book{biber2009register,
  title={Register, Genre, and Style},
  author={Biber, D. and Conrad, S.},
  isbn={9781139483100},
  series={Cambridge Textbooks in Linguistics},
  url={https://books.google.com/books?id=gaEgAwAAQBAJ},
  year={2009},
  publisher={Cambridge University Press}
}

@article{qwen3,
    title={Qwen3 Technical Report}, 
    author={Qwen, Team},
  year={2025},
  eprint={2505.09388},
  archivePrefix={arXiv},
  primaryClass={cs.CL},
  url={https://arxiv.org/abs/2505.09388}, 
}
\bibliographystyle{acl_natbib}

\appendix
\section{Additional Experimental Details}

\subsection{Adaptive Steering Algorithms}

We present the logic for the two phases of our adaptive search: Calibration (Alg.~\ref{alg:calibration}) and Adaptation (Alg.~\ref{alg:adaptation}).

\begin{algorithm}[h]
\small
\caption{Phase I: Calibration (Binary Search)}
\label{alg:calibration}
\begin{algorithmic}[1]
\REQUIRE Model $\mathcal{M}$, Sample $x_0$, Vector $s$, Bounds $[\alpha_{lo}, \alpha_{hi}]$
\STATE \textbf{Function} $\textsc{Score}(\alpha, x)$:
\STATE \quad $y \leftarrow \mathcal{M}.\text{generate}(x, \text{steer}=\alpha \cdot s)$
\STATE \quad \textbf{if} $\text{IsDegenerated}(y)$ \textbf{return} $-\infty$
\STATE \quad $P_{coll} \leftarrow \text{SoftPenalty}(y)$ \COMMENT{Eq. \ref{eq:penalty}}
\STATE \quad \textbf{return} $\alpha \cdot |y| / (1 + P_{coll})$
\STATE \textbf{End Function}

\STATE $\alpha_{best} \leftarrow \alpha_{lo}$
\WHILE{$\alpha_{hi} - \alpha_{lo} > \epsilon$}
    \STATE $\alpha_{mid} \leftarrow (\alpha_{lo} + \alpha_{hi}) / 2$
    \STATE \textbf{if} $\textsc{Score}(\alpha_{mid}, x_0) > -\infty$ \textbf{then}
    \STATE \quad $\alpha_{best} \leftarrow \alpha_{mid}; \alpha_{lo} \leftarrow \alpha_{mid}$
    \STATE \textbf{else} $\alpha_{hi} \leftarrow \alpha_{mid}$
\ENDWHILE
\RETURN $\alpha^*_{base} \leftarrow \alpha_{best}$
\end{algorithmic}
\end{algorithm}

\begin{algorithm}[h]
\small
\caption{Phase II: Adaptation (Heuristic Probing)}
\label{alg:adaptation}
\begin{algorithmic}[1]
\REQUIRE Samples $\{x_i\}_{i=1}^N$, History $\mathcal{H}=\{\alpha^*_{base}\}$, Step $\delta$
\FOR{$i = 1$ \textbf{to} $N$}
    \STATE $\bar{\alpha} \leftarrow \text{Mean}(\mathcal{H})$ \COMMENT{Adaptive Anchor}
    \STATE $\alpha \leftarrow \bar{\alpha}$
    
    \IF{$\textsc{Score}(\bar{\alpha}, x_i) > -\infty$} 
        \STATE \COMMENT{Valid Anchor: Probe Upward}
        \WHILE{$\alpha + \delta \leq \alpha_{max}$ \AND $\textsc{Score}(\alpha+\delta, x_i) > -\infty$}
             \STATE $\alpha \leftarrow \alpha + \delta$
        \ENDWHILE
    \ELSE 
        \STATE \COMMENT{Invalid Anchor: Recover Downward}
        \WHILE{$\alpha - \delta \geq \alpha_{min}$ \AND $\textsc{Score}(\alpha, x_i) = -\infty$}
             \STATE $\alpha \leftarrow \alpha - \delta$
        \ENDWHILE
    \ENDIF
    
    \STATE $\alpha^*_i \leftarrow \textsc{GoldenSearch}(\text{range } [\alpha-\delta, \alpha+\delta], x_i)$
    \STATE Append $\alpha^*_i$ to $\mathcal{H}$
\ENDFOR
\RETURN $\{\alpha^*_i\}_{i=1}^N$
\end{algorithmic}
\end{algorithm}

\subsection{Implementation Details}
\label{app:implementation}

We provide the configuration choices needed to reproduce \AThreeS\ end-to-end.

\paragraph{Backbones and decoding.}
We use \texttt{Llama-3.2-3B-Instruct}~\citep{grattafiori2024llama3herdmodels} and \texttt{Qwen3-4B}~\citep{qwen3} as frozen backbones. Generation uses greedy decoding (temperature $0$, top-$p{=}1$), with a maximum of 256 new tokens for MUD/LaMP/GYAFC and 512 for LongLaMP. Steering is applied during both prefill and decoding. All generation, including the contrastive paraphrases used to construct aspect directions, uses the same backbone and decoding configuration.

\paragraph{Direction extraction.}
For each aspect $d$, the target exemplar serves as the positive text $y_d^+$, while $K{=}4$ contrastive variants $\{y_{d,k}^-\}_{k=1}^{K}$ are generated from the exemplar using the aspect-$d$ contrastive prompt (Appendix~\ref{app:aspect_prompts}). We read the residual-stream states at layer $\ell$ and mean-pool over all tokens of each text. The aspect direction is obtained by averaging the differences
$h_\ell(y_d^+) - h_\ell(y_{d,k}^-)$ over the $K$ sampled variants. We then merge the resulting aspect directions using PCB with a keep ratio of $r{=}0.5$. We steer at layers $\{8{:}18\}$ for Llama and $\{11{:}31\}$ for Qwen, applying PCB independently per layer with shared hyperparameters.

\paragraph{PCB and Hybrid Search.} PCB uses a top-$r{=}0.5$ keep ratio on the per-coordinate importance scores before min-max normalization. The calibration phase performs a binary feasibility search over $\alpha \in [0,1]$ with tolerance $\epsilon{=}0.01$. During adaptation, we use step $\delta{=}0.05$ for upward probing or downward recovery, followed by a golden-section search over the resulting local interval. A candidate is hard-rejected if any of: $P(\text{EOS}|y){<}0.1$ at truncation, suffix-repeat ratio ${>}1.0$, or $\mathcal{P}_{\text{collapse}} {>}5.0$. A single scalar $\alpha$ is shared across layers; per-coordinate competition is absorbed into the PCB weights.

\subsection{Dataset Construction Details}
\label{app:datasets}

\paragraph{Reddit Authorship (MUD).}
We use the MUD dataset~\citep{khan2021deepmetriclearningapproach} to evaluate few-shot user imitation. Following \citet{patel2024lowresourceauthorshipstyletransfer, horvitz-etal-2024-tinystyler, yang2025steeringlargelanguagemodels}, we sample 15 source and 15 target authors from the test split. We generate evaluation pairs by matching every individual source text (16 per author) against every target author, represented by a concatenated exemplar, yielding $16 \times 15 \times 15 = 3{,}600$ evaluation pairs.


\paragraph{Twitter Authorship (LaMP).}
We adapt the Tweet Paraphrasing task from the LaMP benchmark~\citep{salemi-etal-2024-lamp} for unsupervised authorship style transfer. Given a source tweet and a distinct target user profile, we construct the target exemplar by randomly selecting and concatenating $K{=}9$ short texts from the target user’s historical posts, resulting in 1{,}496 evaluation samples.

\paragraph{Long-Form Authorship (LongLaMP).}
We utilize the Topic Writing task from the LongLaMP benchmark~\citep{kumar2024longlampbenchmarkpersonalizedlongform}, reformulated as an authorship transfer problem by randomly pairing distinct authors from the pool of 2{,}452 users. For each pair, a text written by the source author serves as the content input, and a separate writing from the target author’s history is used as the style exemplar. Due to the computational cost of long-form generation, we evaluate on a randomly sampled subset of 100 test instances.

\subsection{Detailed Evaluation Metrics}
\label{app:evaluation}

\paragraph{Style Matching Metrics.}
We evaluate stylistic alignment using three embedding-based models designed to capture authorial or stylistic similarity while minimizing topical overlap.

\begin{itemize}
    \item \textbf{LUAR}~\citep{rivera-soto-etal-2021-learning} produces dense representations optimized for authorship verification. Higher cosine similarity between the generated text and the target exemplar indicates stronger stylistic alignment.

    \item \textbf{StyleCAV}~\citep{wegmann-etal-2022-author} is trained to disentangle writing style from topic using contrastive objectives on Reddit data. It is commonly used to evaluate fine-grained stylistic similarity independent of content.

    \item \textbf{StyleDistance}~\citep{patel-etal-2025-styledistance} learns content-independent style representations from synthetic near-paraphrase pairs, explicitly minimizing content leakage.
\end{itemize}

For all three metrics, we report \emph{Towards} and \emph{Away} scores based on cosine similarity:
\begin{equation}
\begin{aligned}
    \text{Towards} &= \frac{1 + \cos(E(y), E(t))}{2}, \\
    \text{Away} &= \frac{1 - \cos(E(y), E(x))}{2},
\end{aligned}
\end{equation}
where $E(\cdot)$ denotes the embedding of the generated output $y$, the target exemplar $t$, or the source input $x$. The Towards score measures stylistic similarity to the target exemplar, while the Away score measures stylistic divergence from the source text.

\paragraph{Meaning Preservation Metrics.}
We assess semantic fidelity using two complementary metrics:

\begin{itemize}
\item \textbf{MIS}~\citep{babakov-etal-2022-large}: A Natural Language Inference (NLI) based metric measuring bidirectional entailment. \item \textbf{SBERT}~\citep{reimers-gurevych-2019-sentence}: Cosine similarity of sentence embeddings. 
\end{itemize}

\paragraph{Target Overlap.}
To detect direct copying of the target exemplar, we compute ROUGE-1, ROUGE-2, and ROUGE-L scores between the generated output and the target text. Lower ROUGE scores indicate that stylistic transfer is achieved without reusing target content.

\paragraph{LLM-Based Evaluation.}
Automatic metrics are complemented by an LLM-as-a-judge protocol. The LLM-as-a-judge evaluation follows an analogous pairwise comparison protocol using GPT-4.1, with the full prompt provided in Appendix~\ref{app:llm_prompts}.

\subsection{Full Main Results}
\label{app:full_main_results}

Table~\ref{tab:main_results_full} reports the complete metric breakdown summarized in the main paper (Tables~\ref{tab:main_results} and~\ref{tab:gyafc_results}).
It contains the Away scores for all three style-fidelity embeddings, the MIS meaning-preservation metric, and ROUGE-1/2 in addition to ROUGE-L. We follow the same Bold/underline convention as in the main tables; \textit{Simple} scores are grayed where they reflect copying of the target exemplar.

\begin{table*}[t]
\centering
\scriptsize
\renewcommand{\arraystretch}{0.92}
\setlength{\tabcolsep}{2.4pt}
\resizebox{\textwidth}{!}{%
\begin{tabular}{llc cccccc cc ccc}
\toprule
\multirow{2}{*}{\textbf{Dataset}} &
\multirow{2}{*}{\textbf{Method}} &
\multirow{2}{*}{\textbf{Backbone}} &
\multicolumn{2}{c}{\textbf{LUAR} ($\uparrow$)} &
\multicolumn{2}{c}{\textbf{StyleCAV} ($\uparrow$)} &
\multicolumn{2}{c}{\textbf{StyleDistance} ($\uparrow$)} &
\multicolumn{2}{c}{\textbf{Meaning} ($\uparrow$)} &
\multicolumn{3}{c}{\textbf{Target Overlap} ($\downarrow$)} \\
\cmidrule(lr){4-5}
\cmidrule(lr){6-7}
\cmidrule(lr){8-9}
\cmidrule(lr){10-11}
\cmidrule(lr){12-14}
& & & Towards & Away & Towards & Away & Towards & Away
& MIS & SBERT & R-1 & R-2 & R-L \\
\midrule

\multirow{15}{*}{\shortstack[l]{\textbf{MUD}\\(Reddit)}}
& \multirow{2}{*}{Simple}
& Llama
& \textcolor{gray}{0.849} & \textcolor{gray}{0.313}
& \textcolor{gray}{0.833} & \textcolor{gray}{0.384}
& \textcolor{gray}{0.937} & \textcolor{gray}{0.128}
& \textcolor{gray}{0.205} & \textcolor{gray}{0.293}
& \textcolor{gray}{0.606} & \textcolor{gray}{0.571}
& \textcolor{gray}{0.591} \\
& & Qwen
& \textcolor{gray}{0.770} & \textcolor{gray}{0.207}
& \textcolor{gray}{0.737} & \textcolor{gray}{0.302}
& \textcolor{gray}{0.894} & \textcolor{gray}{0.099}
& \textcolor{gray}{0.457} & \textcolor{gray}{0.607}
& \textcolor{gray}{0.435} & \textcolor{gray}{0.382}
& \textcolor{gray}{0.411} \\

& \multirow{2}{*}{STYLL}
& Llama
& 0.636 & \textbf{0.317}
& 0.545 & 0.365
& 0.852 & 0.127
& 0.330 & 0.541
& 0.157 & 0.034 & 0.092 \\
& & Qwen
& 0.651 & 0.262
& 0.527 & \underline{0.390}
& 0.835 & \underline{0.130}
& 0.710 & 0.699
& 0.152 & 0.025 & 0.084 \\

& \multirow{2}{*}{RG}
& Llama
& 0.634 & 0.278
& 0.583 & 0.333
& 0.856 & 0.114
& 0.529 & 0.635
& 0.137 & 0.026 & 0.080 \\
& & Qwen
& 0.653 & 0.201
& 0.533 & 0.312
& 0.837 & 0.105
& \underline{0.774} & \underline{0.796}
& 0.136 & 0.018 & 0.074 \\

& \multirow{2}{*}{Aspect Prompt}
& Llama
& 0.654 & \underline{0.304}
& 0.558 & 0.369
& 0.850 & 0.128
& 0.438 & 0.597
& 0.215 & 0.042 & 0.111 \\
& & Qwen
& 0.653 & 0.265
& 0.478 & \textbf{0.411}
& 0.826 & \textbf{0.135}
& 0.554 & 0.662
& 0.226 & 0.037 & 0.113 \\

\cmidrule(lr){2-14}

& \multirow{2}{*}{Konen et al.}
& Llama
& 0.637 & 0.258
& 0.502 & 0.354
& 0.837 & 0.128
& 0.461 & 0.573
& 0.110 & 0.011 & 0.065 \\
& & Qwen
& 0.651 & 0.208
& 0.512 & 0.326
& 0.846 & 0.117
& 0.668 & 0.729
& 0.095 & \underline{0.009} & 0.057 \\

& \multirow{2}{*}{Global Steering}
& Llama
& 0.626 & 0.232
& 0.552 & 0.306
& 0.856 & 0.104
& 0.558 & 0.712
& 0.117 & 0.012 & 0.065 \\
& & Qwen
& \underline{0.660} & 0.180
& 0.655 & 0.338
& 0.862 & 0.120
& 0.732 & 0.770
& \underline{0.067} & \textbf{0.005} & 0.045 \\

& TinyStyler
& --
& 0.648 & 0.237
& \textbf{0.788} & 0.369
& \textbf{0.885} & 0.119
& 0.637 & 0.735
& 0.104 & \underline{0.009} & 0.068 \\

& \multirow{2}{*}{\textbf{\AThreeS}}
& Llama
& 0.656 & 0.232
& 0.676 & 0.378
& 0.872 & 0.126
& 0.575 & 0.681
& \textbf{0.058} & \textbf{0.005} & \textbf{0.039} \\
& & Qwen
& \textbf{0.665} & 0.186
& \underline{0.678} & 0.332
& \underline{0.873} & 0.118
& \textbf{0.781} & \textbf{0.821}
& 0.073 & \textbf{0.005} & \underline{0.044} \\

\midrule

\multirow{15}{*}{\shortstack[l]{\textbf{LaMP}\\(Tweet)}}
& \multirow{2}{*}{Simple}
& Llama
& \textcolor{gray}{0.826} & \textcolor{gray}{0.263}
& \textcolor{gray}{0.821} & \textcolor{gray}{0.527}
& \textcolor{gray}{0.918} & \textcolor{gray}{0.189}
& \textcolor{gray}{0.355} & \textcolor{gray}{0.472}
& \textcolor{gray}{0.421} & \textcolor{gray}{0.362}
& \textcolor{gray}{0.406} \\
& & Qwen
& \textcolor{gray}{0.814} & \textcolor{gray}{0.173}
& \textcolor{gray}{0.679} & \textcolor{gray}{0.299}
& \textcolor{gray}{0.866} & \textcolor{gray}{0.115}
& \textcolor{gray}{0.558} & \textcolor{gray}{0.660}
& \textcolor{gray}{0.450} & \textcolor{gray}{0.393}
& \textcolor{gray}{0.428} \\

& \multirow{2}{*}{STYLL}
& Llama
& 0.667 & \textbf{0.273}
& 0.563 & 0.356
& 0.825 & 0.137
& 0.528 & 0.615
& 0.119 & 0.022 & 0.080 \\
& & Qwen
& 0.693 & 0.208
& 0.434 & 0.239
& 0.787 & 0.089
& 0.812 & 0.738
& 0.137 & 0.022 & 0.087 \\

& \multirow{2}{*}{RG}
& Llama
& 0.690 & 0.245
& 0.656 & 0.449
& 0.852 & 0.161
& 0.706 & 0.689
& 0.101 & 0.013 & 0.066 \\
& & Qwen
& 0.712 & 0.173
& 0.431 & 0.231
& 0.790 & 0.088
& \textbf{0.857} & \textbf{0.795}
& 0.143 & 0.020 & 0.088 \\

& \multirow{2}{*}{Aspect Prompt}
& Llama
& 0.702 & 0.257
& 0.634 & 0.417
& 0.844 & 0.156
& 0.640 & 0.678
& 0.161 & 0.024 & 0.094 \\
& & Qwen
& \underline{0.741} & 0.211
& 0.415 & 0.306
& 0.778 & 0.105
& 0.771 & 0.708
& 0.194 & 0.028 & 0.110 \\

\cmidrule(lr){2-14}

& \multirow{2}{*}{Konen et al.}
& Llama
& 0.696 & 0.232
& 0.487 & 0.332
& 0.803 & 0.108
& 0.485 & 0.582
& 0.124 & 0.013 & 0.080 \\
& & Qwen
& 0.727 & 0.191
& 0.503 & 0.302
& 0.823 & 0.126
& 0.704 & 0.714
& 0.101 & 0.011 & 0.068 \\

& \multirow{2}{*}{Global Steering}
& Llama
& 0.685 & 0.220
& 0.567 & 0.329
& 0.833 & 0.108
& 0.647 & 0.704
& 0.130 & 0.014 & 0.080 \\
& & Qwen
& 0.740 & 0.213
& 0.737 & 0.519
& 0.881 & 0.184
& 0.811 & 0.707
& 0.074 & \underline{0.007} & 0.052 \\

& TinyStyler
& --
& 0.727 & 0.232
& \textbf{0.811} & 0.475
& 0.871 & 0.154
& 0.706 & 0.746
& 0.139 & 0.017 & 0.095 \\

& \multirow{2}{*}{\textbf{\AThreeS}}
& Llama
& 0.729 & \underline{0.262}
& \underline{0.763} & \textbf{0.585}
& \underline{0.886} & \underline{0.186}
& 0.479 & 0.551
& \textbf{0.051} & \textbf{0.005} & \textbf{0.037} \\
& & Qwen
& \textbf{0.742} & 0.218
& 0.756 & \underline{0.528}
& \textbf{0.896} & \textbf{0.191}
& \underline{0.841} & \underline{0.748}
& \underline{0.072} & \underline{0.007} & \underline{0.050} \\

\midrule

\multirow{15}{*}{\shortstack[l]{\textbf{LongLaMP}\\(Writing)}}
& \multirow{2}{*}{Simple}
& Llama
& \textcolor{gray}{0.811} & \textcolor{gray}{0.274}
& \textcolor{gray}{0.773} & \textcolor{gray}{0.264}
& \textcolor{gray}{0.919} & \textcolor{gray}{0.098}
& \textcolor{gray}{0.150} & \textcolor{gray}{0.304}
& \textcolor{gray}{0.620} & \textcolor{gray}{0.524}
& \textcolor{gray}{0.570} \\
& & Qwen
& \textcolor{gray}{0.767} & \textcolor{gray}{0.173}
& \textcolor{gray}{0.747} & \textcolor{gray}{0.211}
& \textcolor{gray}{0.902} & \textcolor{gray}{0.082}
& \textcolor{gray}{0.314} & \textcolor{gray}{0.649}
& \textcolor{gray}{0.432} & \textcolor{gray}{0.321}
& \textcolor{gray}{0.370} \\

& \multirow{2}{*}{STYLL}
& Llama
& 0.619 & \textbf{0.330}
& 0.619 & \underline{0.365}
& 0.858 & \textbf{0.140}
& 0.162 & 0.658
& 0.220 & 0.024 & 0.116 \\
& & Qwen
& 0.640 & 0.251
& 0.609 & 0.357
& 0.858 & 0.128
& 0.551 & 0.842
& 0.228 & 0.031 & 0.120 \\

& \multirow{2}{*}{RG}
& Llama
& 0.621 & 0.273
& 0.627 & 0.336
& 0.860 & 0.124
& 0.417 & 0.783
& 0.207 & 0.019 & 0.104 \\
& & Qwen
& 0.649 & 0.172
& 0.612 & 0.310
& 0.868 & 0.100
& \textbf{0.585} & \textbf{0.926}
& 0.217 & 0.020 & 0.104 \\

& \multirow{2}{*}{Aspect Prompt}
& Llama
& 0.624 & 0.269
& 0.638 & 0.337
& 0.864 & 0.120
& 0.395 & 0.784
& 0.220 & 0.022 & 0.109 \\
& & Qwen
& 0.652 & 0.186
& 0.601 & 0.362
& 0.862 & 0.116
& \underline{0.566} & \underline{0.916}
& 0.214 & 0.021 & 0.103 \\

\cmidrule(lr){2-14}

& \multirow{2}{*}{Konen et al.}
& Llama
& 0.660 & \underline{0.297}
& 0.646 & \textbf{0.370}
& 0.865 & \underline{0.137}
& 0.300 & 0.654
& 0.232 & 0.026 & 0.123 \\
& & Qwen
& \underline{0.665} & 0.220
& 0.648 & 0.354
& 0.869 & 0.124
& 0.444 & 0.813
& 0.230 & 0.022 & 0.114 \\

& \multirow{2}{*}{Global Steering}
& Llama
& 0.658 & 0.279
& 0.621 & \textbf{0.370}
& 0.869 & 0.127
& 0.245 & 0.551
& \underline{0.169} & \underline{0.017} & \underline{0.088} \\
& & Qwen
& 0.658 & 0.227
& 0.701 & 0.324
& \underline{0.877} & 0.106
& 0.484 & 0.836
& 0.215 & 0.021 & 0.103 \\

& TinyStyler
& --
& 0.628 & 0.255
& \underline{0.714} & 0.323
& 0.875 & 0.122
& 0.252 & 0.703
& \textbf{0.133} & \textbf{0.010} & \textbf{0.083} \\

& \multirow{2}{*}{\textbf{\AThreeS}}
& Llama
& 0.663 & 0.228
& 0.640 & 0.348
& \underline{0.877} & 0.108
& 0.408 & 0.788
& 0.215 & 0.022 & 0.114 \\
& & Qwen
& \textbf{0.674} & 0.204
& \textbf{0.715} & 0.324
& \textbf{0.898} & 0.102
& 0.521 & 0.863
& 0.214 & 0.020 & 0.103 \\

\bottomrule
\end{tabular}}
\caption{Full results across the three authorship-transfer benchmarks using Llama and Qwen backbones. Towards and Away measure style fidelity; MIS and SBERT measure meaning preservation; and ROUGE-1/2/L measure target-exemplar overlap. \textbf{Bold} and \underline{underline} indicate the best and second-best non-grayed values per dataset. Simple is grayed because its high target overlap indicates copying.}
\label{tab:main_results_full}
\end{table*}

\subsection{Formality Transfer as a Contrast}
\label{app:gyafc}

We use a formality transfer task as a control condition: when target style reduces to a single coarse attribute (formal vs. informal), the hypothesis that aspect decomposition captures fine-grained authorial style predicts that \AThreeS's advantage over simpler baselines should diminish.

We use the  Entertainment \& Music (EM) of the \textbf{Yahoo Answers Authorship (GYAFC)} dataset as a benchmark \citep{rao-tetreault-2018-dear}, and reframe the task as style transfer based on a style examplar. We apply our method in both Informal$\to$Formal and Formal$\to$Informal directions. Target exemplars are constructed by concatenating 16 randomly selected sentences from the training split corresponding to the target formality.

The results confirm the hypothesis (Table~\ref{tab:gyafc_results}), with \AThreeS~showing smaller improvements over Global Steering and RG baselines than on multi-aspect tasks. TinyStyler leads the three style metrics, but \AThreeS~(Qwen) has the highest SBERT in both directions (0.775 and 0.732). The preference evaluation using GPT-4.1 confirms that \AThreeS~and TinyStyler are effectively tied with 50\% of preferences for \AThreeS~and 45\% for TinyStyler.

These results further delimit the regime where compositional aspect decomposition is expected to help: when authorship style genuinely combines multiple rhetorical dimensions rather than collapsing to a single transferable attribute. In this single-attribute limit, \AThreeS~still outperforms prompting baselines while matching the dedicated trained model, suggesting that activation-space style representations remain useful.

\begin{table}[t]
\centering
\footnotesize
\renewcommand{\arraystretch}{0.95}
\setlength{\tabcolsep}{2.6pt}
\resizebox{\columnwidth}{!}{%
\begin{tabular}{ll cccccc}
\toprule
\textbf{Method} & \textbf{Backbone} & \textbf{LUAR} & \textbf{CAV} & \textbf{SD} & \textbf{SBERT} & \textbf{R-L} & \textbf{Len} \\
& & ($\uparrow$) & ($\uparrow$) & ($\uparrow$) & ($\uparrow$) & ($\downarrow$) & \\
\midrule
\multicolumn{8}{l}{\textit{GYAFC (Formal $\to$ Informal)}} \\
Simple          & Llama & \textcolor{gray}{0.845} & \textcolor{gray}{0.836} & \textcolor{gray}{0.941} & \textcolor{gray}{0.305} & \textcolor{gray}{0.566} & \textcolor{gray}{12.77} \\
STYLL           & Llama & 0.670 & 0.555 & 0.857 & 0.437 & 0.120 & 8.24 \\
RG              & Llama & 0.651 & 0.473 & 0.835 & 0.601 & 0.085 & 4.67 \\
Aspect Prompt & Llama & 0.681 & 0.421 & 0.822 & 0.533 & 0.138 & 11.22 \\
\cmidrule(lr){1-8}
Global Steering & Llama & 0.644 & 0.457 & 0.821 & \underline{0.726} & 0.061 & 2.91 \\
TinyStyler      & -   & \textbf{0.702} & \textbf{0.872} & \textbf{0.919} & 0.713 & 0.094 & 5.04 \\
\rowcolor{lightgray!25}
\textbf{\AThreeS} & Llama & 0.688 & \underline{0.772} & \underline{0.896} & 0.604 & \textbf{0.032} & 1.02 \\
\rowcolor{lightgray!25}
\textbf{\AThreeS} & Qwen  & \underline{0.690} & 0.707 & 0.893 & \textbf{0.775} & \underline{0.036} & 1.23 \\
\midrule

\multicolumn{8}{l}{\textit{GYAFC (Informal $\to$ Formal)}} \\
Simple          & Llama & \textcolor{gray}{0.826} & \textcolor{gray}{0.794} & \textcolor{gray}{0.918} & \textcolor{gray}{0.349} & \textcolor{gray}{0.533} & \textcolor{gray}{13.66} \\
STYLL           & Llama & 0.653 & 0.594 & 0.857 & 0.463 & 0.149 & 12.88 \\
RG              & Llama & 0.620 & 0.623 & 0.860 & 0.600 & 0.086 & 5.35 \\
Aspect Prompt & Llama & 0.661 & \underline{0.660} & \underline{0.874} & 0.516 & 0.151 & 12.57 \\
\cmidrule(lr){1-8}
Global Steering & Llama & 0.638 & 0.563 & 0.871 & 0.588 & 0.083 & 5.07 \\
TinyStyler      & -   & \textbf{0.727} & \textbf{0.880} & \textbf{0.926} & \underline{0.685} & 0.106 & 4.91 \\
\rowcolor{lightgray!25}
\textbf{\AThreeS} & Llama & 0.682 & 0.568 & 0.871 & 0.497 & \underline{0.057} & 1.82 \\
\rowcolor{lightgray!25}
\textbf{\AThreeS} & Qwen  & \underline{0.692} & 0.627 & 0.854 & \textbf{0.732} & \textbf{0.052} & 2.28 \\
\bottomrule
\end{tabular}}
\caption{Single-attribute formality transfer (GYAFC). Same metrics and ranking conventions as Table~\ref{tab:main_results}.}
\label{tab:gyafc_results}
\end{table}

\subsection{Layer-wise Accuracy by Dimension}\label{app:layer_analysis}

\begin{figure}[h]
    \centering
    \includegraphics[width=\columnwidth]{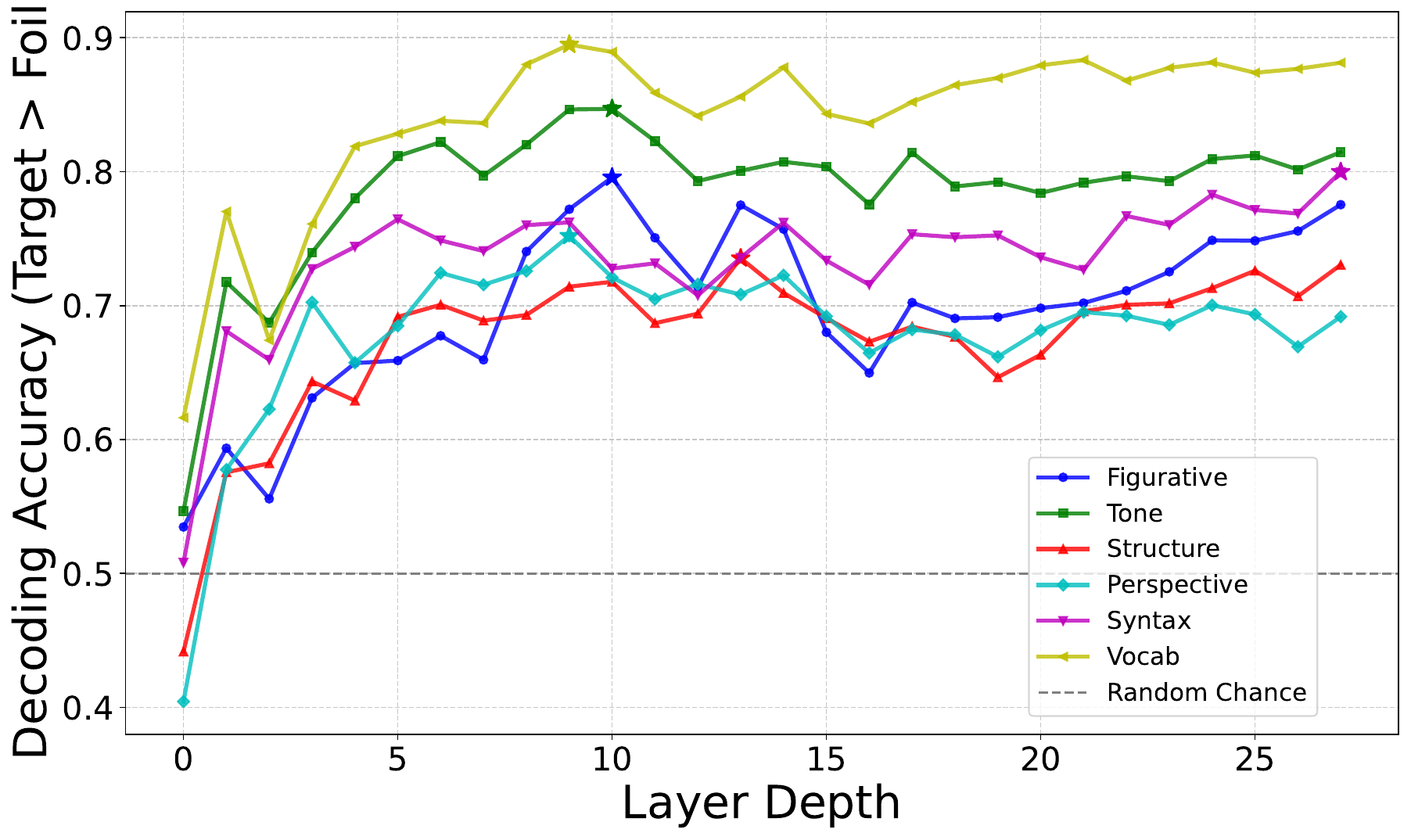}
    \caption{Layer-wise classification accuracy for different stylistic dimensions. The y-axis represents the probability that the projection of the hidden state onto the learned direction is higher for the target author than the foil. We observe that high-level concepts like \textit{Perspective} and \textit{Tone} peak in the middle-to-late layers (Layers 10--20).}
    \label{fig:style_layers}
\end{figure}

Figure~\ref{fig:style_layers} illustrates the decoding accuracy of the representation vectors for six different stylistic dimensions (Figurative, Tone, Structure, Perspective, Syntax, Vocabulary). The accuracy is measured by the ability of the vector to distinguish between the target author's text and a foil author's text in a held-out development set.

As shown in Figure~\ref{fig:style_layers}, most stylistic dimensions achieve peak separability in the middle layers of the network. This supports our decision to intervene at these specific layers during the generation process.

\subsection{Full Stylistic Dimensions \& Aggregation Ablation Results}
\label{app:full_aspect_analysis}

This appendix reports the complete ablation results for all stylistic dimensions,
aggregation strategies, and evaluation metrics (Table~\ref{tab:full_agg_ablation}). While the main paper focuses on
StyleCAV and SBERT for clarity and space constraints, we include all metrics here
for completeness and reproducibility.

\begin{table*}[h]
\centering
\small
\setlength{\tabcolsep}{4.5pt}
\begin{tabular}{lcccccc}
\toprule
& \multicolumn{3}{c}{\textbf{Style Fidelity} ($\uparrow$)} & \multicolumn{3}{c}{\textbf{Meaning Preservation} ($\uparrow$)} \\
\cmidrule(lr){2-4} \cmidrule(lr){5-7}
\textbf{Configuration} & \textbf{LUAR} & \textbf{StyleCAV} & \textbf{StyleDistance} & \textbf{SBERT} & \textbf{MIS} & \textbf{METEOR} \\
\midrule
\multicolumn{7}{l}{\textit{Baselines \& Individual Dimensions}} \\
Neutral (No Aspect) & 0.227 & 0.055 & 0.710 & 0.799 & 0.670 & 0.602 \\
Figurative & 0.288 & 0.463 & 0.761 & 0.685 & 0.560 & 0.422 \\
Tone & 0.294 & 0.427 & 0.744 & 0.665 & 0.597 & 0.399 \\
Perspective & 0.297 & 0.403 & 0.755 & 0.713 & 0.677 & 0.417 \\
Structure & 0.275 & 0.399 & 0.757 & 0.750 & 0.673 & 0.489 \\
Syntax & 0.251 & 0.333 & 0.741 & 0.727 & 0.719 & 0.518 \\
Vocabulary & 0.240 & 0.310 & 0.739 & 0.724 & 0.620 & 0.415 \\
Surface & 0.194 & 0.103 & 0.666 & 0.692 & 0.535 & 0.400 \\
\midrule
\multicolumn{7}{l}{\textit{PCB-Merging}} \\
Figurative + Tone & 0.301 & 0.483 & 0.758 & 0.663 & 0.607 & 0.376 \\
Figurative + Tone + Perspective & 0.295 & 0.461 & 0.756 & 0.665 & 0.618 & 0.376 \\
Figurative + Tone + Perspective + Structure & 0.304 & 0.485 & 0.753 & 0.694 & 0.643 & 0.368 \\
All Aspects & 0.291 & 0.388 & 0.747 & 0.713 & 0.642 & 0.443 \\
\midrule
\multicolumn{7}{l}{\textit{Mean Averaging}} \\
Figurative + Tone & 0.312 & 0.446 & 0.752 & 0.627 & 0.521 & 0.321 \\
Figurative + Tone + Perspective & 0.297 & 0.462 & 0.753 & 0.656 & 0.589 & 0.362 \\
Figurative + Tone + Perspective + Structure & 0.297 & 0.442 & 0.756 & 0.659 & 0.613 & 0.369 \\
All Aspects & 0.289 & 0.396 & 0.742 & 0.675 & 0.645 & 0.393 \\
\midrule
\multicolumn{7}{l}{\textit{Median Aggregation}} \\
Figurative + Tone & 0.308 & 0.456 & 0.754 & 0.642 & 0.656 & 0.363 \\
Figurative + Tone + Perspective & 0.292 & 0.428 & 0.753 & 0.674 & 0.594 & 0.389 \\
Figurative + Tone + Perspective + Structure & 0.297 & 0.459 & 0.758 & 0.713 & 0.632 & 0.400 \\
All Aspects & 0.287 & 0.422 & 0.751 & 0.686 & 0.601 & 0.385 \\
\midrule
\multicolumn{7}{l}{\textit{TIES-Merging}} \\
Figurative + Tone + Perspective & 0.307 & 0.438 & 0.741 & 0.646 & 0.590 & 0.348 \\
Figurative + Tone + Perspective + Structure & 0.307 & 0.428 & 0.753 & 0.650 & 0.627 & 0.363 \\
All Aspects & 0.281 & 0.390 & 0.746 & 0.675 & 0.653 & 0.420 \\
\bottomrule
\end{tabular}
\caption{Full ablation results for individual style aspects and aggregation strategies. We report Style Fidelity (LUAR, StyleCAV, StyleDistance) and Meaning Preservation (SBERT, MIS, METEOR). Neutral represents standard activation subtraction without aspect-specific prompting.}
\label{tab:full_agg_ablation}
\end{table*}

\subsection{Full Adaptive $\alpha$ Ablation Results}
\label{app:alpha_full}

Table~\ref{tab:ablation_alpha_full} reports the complete results for the adaptive $\alpha$ ablation, including all style and meaning preservation metrics omitted from the main paper for space. All experiments use PCB-Merging to merge the four highest-performing stylistic dimensions (\textit{Figurative, Tone, Perspective, and Structure}).

\begin{table*}[t]
\centering
\small
\setlength{\tabcolsep}{5pt}
\begin{tabular}{lccccccc}
\toprule
\textbf{Configuration} &
\textbf{LUAR} &
\textbf{StyleCAV} &
\textbf{StyleDistance} &
\textbf{SBERT} &
\textbf{MIS} &
\textbf{METEOR} &
\textbf{Time (s)} \\
\midrule
$\alpha = 0.0$ & 0.137 & -0.196 & 0.625 & 0.807 & 0.724 & 0.591 & 0 \\
$\alpha = 0.1$ & 0.118 & -0.220 & 0.636 & 0.863 & 0.910 & 0.863 & 0 \\
$\alpha = 0.2$ & 0.128 & -0.093 & 0.655 & 0.880 & 0.912 & 0.689 & 0 \\
$\alpha = 0.3$ & 0.161 & 0.170 & 0.708 & 0.910 & 0.901 & 0.759 & 0 \\
$\alpha = 0.4$ & 0.199 & 0.307 & 0.741 & 0.874 & 0.827 & 0.729 & 0 \\
$\alpha = 0.5$ & 0.247 & 0.409 & 0.752 & 0.761 & 0.704 & 0.612 & 0 \\
$\alpha = 0.6$ & 0.274 & 0.385 & 0.738 & 0.689 & 0.617 & 0.515 & 0 \\
$\alpha = 0.7$ & 0.312 & 0.344 & 0.715 & 0.600 & 0.560 & 0.412 & 0 \\
$\alpha = 0.8$ & 0.352 & 0.329 & 0.701 & 0.444 & 0.455 & 0.242 & 0 \\
$\alpha = 0.9$ & 0.375 & 0.311 & 0.691 & 0.296 & 0.348 & 0.139 & 0 \\
\midrule
Binary Search & 0.311 & 0.459 & 0.753 & 0.639 & 0.606 & 0.348 & 15.56 \\
Hybrid Search (Ours) &
0.303 &
0.480 &
0.753 &
0.694 &
0.643 &
0.398 &
11.73 \\
\bottomrule
\end{tabular}
\caption{Full results for fixed and adaptive steering coefficients on the MUD development set. Style metrics report Towards scores (higher is better).}
\label{tab:ablation_alpha_full}
\end{table*}

\subsection{Additional Length and Preference Results}
\label{app:length_llm}

We provide additional analyses of output length and direct preferences between \AThreeS\ and TinyStyler. The detailed length-shift statistics below correspond to \AThreeS\ with the Llama backbone; aggregate Qwen length ratios are reported in Table~\ref{tab:main_results}.

\begin{table}[h]
\centering
\small
\setlength{\tabcolsep}{3.5pt}
\begin{tabular}{llccc}
\toprule
\textbf{Dataset} & \textbf{System} & \textbf{Ratio} & \textbf{$>10\%$} & \textbf{Abs. Shift} \\
\midrule
MUD & \textsc{A3S} & 0.876 & 10.97 & 0.415 \\
 & TinyStyler & 4.399 & 69.17 & 3.624 \\
LaMP & \textsc{A3S} & 0.838 & 12.70 & 0.457 \\
 & TinyStyler & 2.325 & 93.85 & 1.339 \\
LongLaMP & \textsc{A3S} & 0.779 & 14.00 & 0.332 \\
 & TinyStyler & 0.315 & 0.00 & 0.685 \\
\bottomrule
\end{tabular}
\caption{Length-shift statistics. Ratio is $\mathrm{len(output)}/\mathrm{len(input)}$ in characters; $>10\%$ is the fraction of outputs more than 10\% longer than the source.}
\label{tab:length_shift}
\end{table}

\paragraph{Human preference on MUD.}
We compare \AThreeS~(Qwen) against TinyStyler using 210 judgments
across 70 MUD instances, with three independent judgments per item.
Annotators select \AThreeS\ in 114 judgments and TinyStyler in 71,
with 25 ties, corresponding to preference rates of 54.3\%, 33.8\%,
and 11.9\%, respectively. After excluding ties, the \AThreeS\ win
rate is 61.6\% (95\% item-level CI: $[53.3\%,69.7\%]$;
$p=0.0019$).

\paragraph{LLM-as-a-judge preference.}
We evaluate both \AThreeS~(Llama) and \AThreeS~(Qwen) against
TinyStyler using GPT-4.1. The Llama rows in
Table~\ref{tab:llm_pairwise} reproduce the original sampled
evaluation, while the Qwen rows cover the full evaluation set
of 5,196 comparisons.

\begin{table}[t]
\centering
\small
\setlength{\tabcolsep}{4pt}
\begin{tabular}{llrrr}
\toprule
\textbf{Dataset} & \textbf{Backbone}
& \multicolumn{3}{c}{\textbf{Preference (\%)}} \\
\cmidrule(lr){3-5}
& & \textbf{\AThreeS} & \textbf{Tie}
& \textbf{TinyStyler} \\
\midrule
MUD
    & Llama & 54.0 &  7.0 & 39.0 \\
    & Qwen  & 66.7 &  6.4 & 26.9 \\
LaMP
    & Llama & 69.0 &  2.0 & 29.0 \\
    & Qwen  & 79.5 &  1.5 & 18.9 \\
LongLaMP
    & Llama & 60.0 & 12.0 & 28.0 \\
    & Qwen  & 92.0 &  2.0 &  6.0 \\
\bottomrule
\end{tabular}
\caption{Pairwise GPT-4.1 preferences comparing \AThreeS\ with
Llama and Qwen backbones against TinyStyler. Percentages include
ties and may not sum to 100 due to rounding.}
\label{tab:llm_pairwise}
\end{table}

GPT-4.1 prefers \AThreeS\ over TinyStyler with both backbones.
For Qwen, the preference is strongest on the out-of-domain
benchmarks LaMP and LongLaMP, but remains substantial on MUD,
TinyStyler's training domain. These results are consistent with
the automatic evaluation, where \AThreeS~(Qwen) exceeds TinyStyler
on both LUAR and SBERT across all three benchmarks. They also align
with the human evaluation on MUD and the original GPT-4.1 evaluation
of the Llama outputs.

\subsection{Aspect Geometry Diagnostics}
\label{app:aspect_geometry}

This appendix provides the full diagnostics underlying the main-body aspect-geometry analysis (Section~\ref{sec:aspect_geometry}).

\begin{figure*}[h]
    \centering
    \begin{minipage}{0.48\textwidth}
        \centering
        \includegraphics[width=\linewidth]{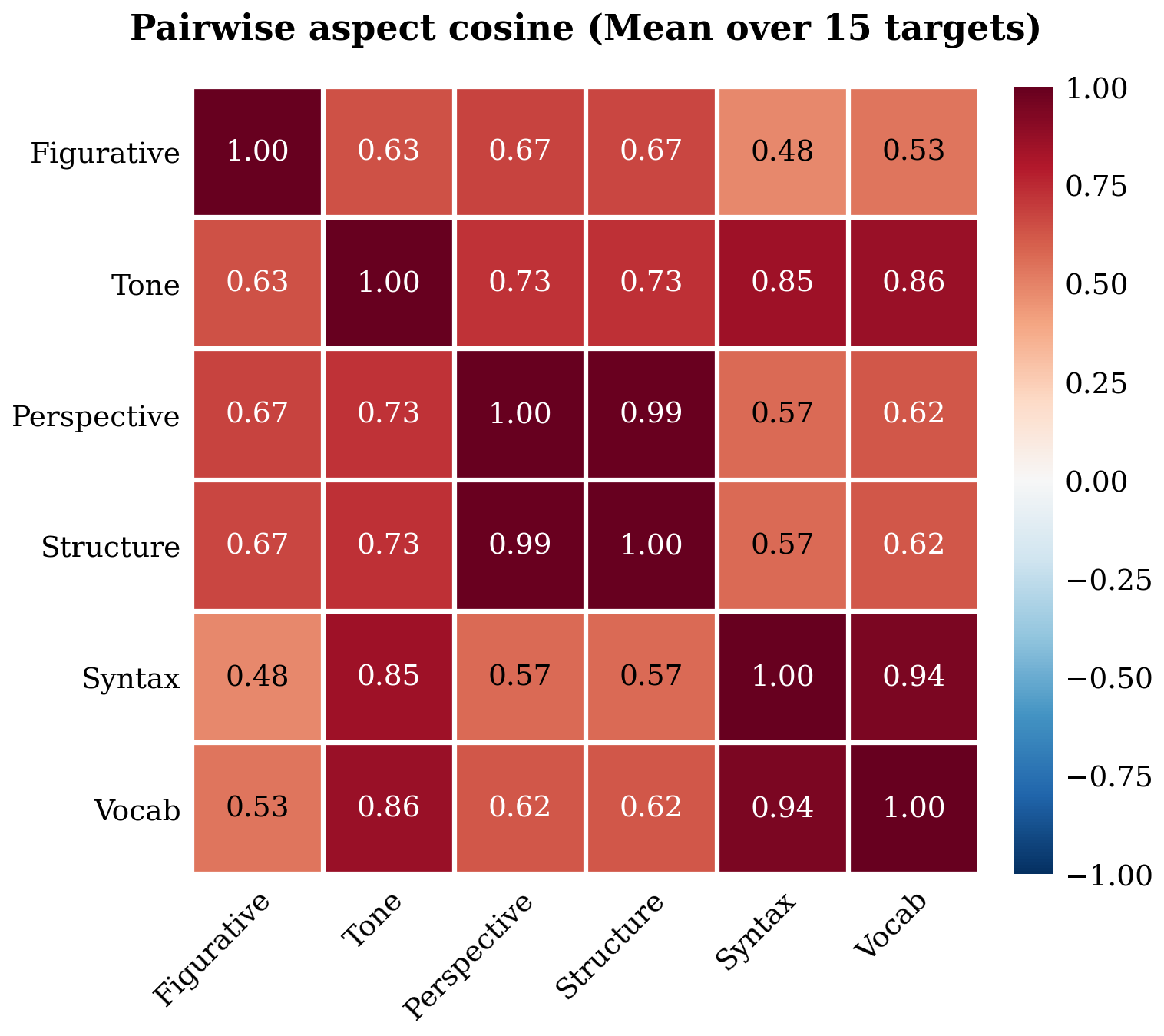}
    \end{minipage}
    \hfill
    \begin{minipage}{0.48\textwidth}
        \centering
        \includegraphics[width=\linewidth]{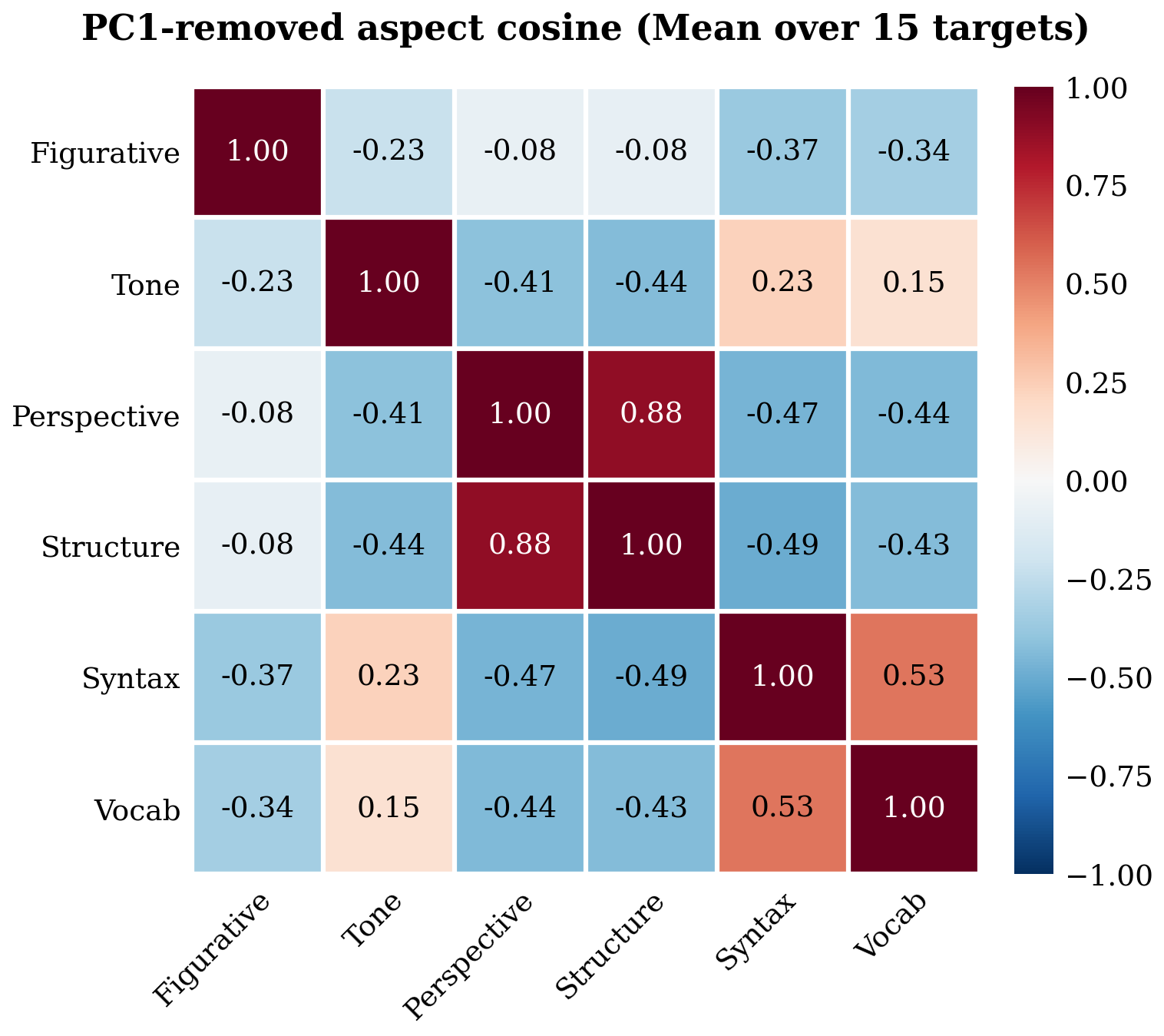}
    \end{minipage}
    \caption{Pairwise cosine similarity between aspect directions before (left) and after removing the first principal component (right). Raw aspect vectors share a strong common authorship component, while PC1-removed residuals expose conflicting aspect-specific corrections.}
    \label{fig:aspect_cosine}
\end{figure*}

\paragraph{Pairwise aspect cosine.}
Figure~\ref{fig:aspect_cosine} shows pairwise cosine similarity between aspect vectors, averaged over 15 target profiles. In the raw activation space, all aspects are positively aligned, indicating a dominant shared authorship component. After removing the first principal component (PC1), the residual directions reveal substantial conflict: 60\% of aspect pairs have negative residual cosine similarity, and discourse-level aspects (\textit{Perspective}, \textit{Structure}) systematically oppose local realization aspects (\textit{Syntax}, \textit{Vocabulary}).

\paragraph{Residuals are non-trivial in magnitude.}
A natural concern is that the residual directions, while geometrically conflicting, might be small enough to be noise. Table~\ref{tab:residual_norm} reports the per-aspect residual-to-total norm ratio. The overall mean is $0.627$, meaning roughly two-thirds of each aspect direction lies outside the shared PC1 backbone. Per-aspect ratios vary: \textit{Tone} and \textit{Vocabulary} are dominated by the shared component (residual ratios $0.36$, $0.37$), while \textit{Figurative}, \textit{Perspective}, and \textit{Structure} carry substantial information outside PC1 (ratios $\geq 0.82$). This is consistent with the cosine heatmap: aspects with smaller residual norms also have weaker residual disagreement.

\paragraph{Decomposing the merger.}
The decomposition results reported in the right panel of Table~\ref{tab:residual_analysis} are reproduced from the MUD development set. Notably, raw PCB over the original aspect vectors ($0.378$ at $\alpha{=}0.5$) exceeds even the explicit PC1 + PCB(residuals) decomposition ($0.339$). We interpret this as evidence that PCB is most effective when it jointly aggregates shared and residual signals, rather than treating them as separate additive components.

\begin{table}[h]
\centering
\small
\setlength{\tabcolsep}{4pt}
\begin{tabular}{lcc}
\toprule
\textbf{Aspect} & \textbf{Mean} & \textbf{Std} \\
\midrule
Figurative   & 0.832 & 0.289 \\
Tone         & 0.359 & 0.110 \\
Perspective  & 0.822 & 0.487 \\
Structure    & 0.831 & 0.474 \\
Syntax       & 0.549 & 0.129 \\
Vocabulary   & 0.370 & 0.134 \\
\midrule
Overall      & 0.627 & -- \\
\bottomrule
\end{tabular}
\caption{Per-aspect residual norm ratio $\|\,s_d - \mathrm{proj}_{\mathrm{PC1}}(s_d)\|/\|s_d\|$, averaged over 15 target profiles. Larger values indicate that more of the aspect direction lies outside the shared PC1 backbone.}
\label{tab:residual_norm}
\end{table}

\subsection{Quality--Compute Comparison}
\label{app:quality_compute}

We compare end-to-end inference cost and output quality on the MUD development set using the Qwen backbone. Table~\ref{tab:quality_compute} reports model calls and prompt/output tokens for each complete method. Prompting costs include target-style analysis, descriptor extraction, and rewriting, whereas activation-based costs include target-profile construction and generation using either fixed-strength steering ($\alpha{=}0.5$) or Hybrid Search. Table~\ref{tab:construction_cost} separately reports the per-profile direction-construction cost.

\begin{table}[h]
\centering
\small
\setlength{\tabcolsep}{3.5pt}
\begin{tabular}{lccc}
\toprule
\textbf{Dataset} & \textbf{\# Profiles} & \textbf{Tokens/Profile} & \textbf{Sec/Profile} \\
\midrule
MUD & 15 & 404{,}812 & 33.00 \\
LongLaMP & 100 & 258{,}328 & 27.84 \\
LaMP & 1{,}496 & 210{,}015 & 27.79 \\
\bottomrule
\end{tabular}
\caption{Cost of constructing \textsc{A3S} directions per target author/profile. Each profile uses 16 generated outputs.}
\label{tab:construction_cost}
\end{table}

\begin{table*}[t]
\centering
\small
\setlength{\tabcolsep}{5pt}
\resizebox{\textwidth}{!}{%
\begin{tabular}{lrrrrr}
\toprule
\textbf{Method} & \textbf{Total Calls} & \textbf{Prompt Tokens} & \textbf{Output Tokens} & \textbf{StyleCAV} $\uparrow$ & \textbf{SBERT} $\uparrow$ \\
\midrule
STYLL & 3.0 & 2{,}808 & 891 & 0.581 & 0.742 \\
STYLL (Reasoning) & 3.0 & 1{,}739 & 1{,}887 & 0.420 & 0.767 \\
RG & 3.0 & 8{,}107 & 2{,}180 & 0.589 & 0.797 \\
RG (Reasoning) & 3.0 & 5{,}938 & 3{,}810 & 0.442 & 0.809 \\
Aspect Prompt & 3.0 & 7{,}306 & 1{,}791 & 0.516 & 0.730 \\
Aspect Prompt (Reasoning) & 3.0 & 6{,}526 & 3{,}299 & 0.389 & 0.744 \\
\midrule
\citet{konen-etal-2024-style} ($\alpha{=}0.5$) & 1.0 & 66 & 151 & 0.511 & 0.485 \\
\citet{konen-etal-2024-style} (Hybrid) & 5.5 & 354 & 351 & 0.522 & 0.720 \\
Global Steering ($\alpha{=}0.5$) & 5.0 & 56{,}583 & 11{,}052 & 0.692 & 0.779 \\
Global Steering (Hybrid) & 11.5 & 57{,}006 & 11{,}351 & 0.712 & 0.801 \\
\AThreeS\ ($\alpha{=}0.5$) & 17.0 & 348{,}014 & 56{,}946 & 0.704 & 0.819 \\
\AThreeS\ (Hybrid) & 23.4 & 348{,}434 & 57{,}300 & \textbf{0.756} & \textbf{0.848} \\
\bottomrule
\end{tabular}}
\caption{End-to-end quality--compute comparison on the MUD development set using the Qwen backbone. Token counts include all stages of each method. The horizontal separator distinguishes prompting-based methods (above) from activation-steering methods (below). Higher StyleCAV and SBERT scores are better.}
\label{tab:quality_compute}
\end{table*}

\paragraph{First-use versus amortized cost.}
The costs reported for \AThreeS in Table~\ref{tab:quality_compute} are driven primarily by the 16 contrastive generations required to construct a new target profile, while Hybrid Search adds relatively little token overhead. Because the resulting directions can be cached and reused, profile construction is a one-time cost that can be amortized over subsequent inputs.

\subsection{Test-Time Reasoning for Prompting Baselines}
\label{app:reasoning_baselines}

STYLL, RG, and Aspect Prompt already use multi-step pipelines that analyze the target style before producing a rewrite. To test whether additional test-time reasoning strengthens these prompting baselines, we also evaluate a reasoning-enabled variant of each method under the same backbone and dataset used in Table~\ref{tab:quality_compute}.

Explicit reasoning substantially increases output-token usage for all three methods, but it does not improve style fidelity. StyleCAV decreases from $0.581$ to $0.420$ for STYLL, from $0.589$ to $0.442$ for RG, and from $0.516$ to $0.389$ for Aspect Prompt. Reasoning yields only modest SBERT improvements ($+0.025$, $+0.012$, and $+0.014$, respectively). These results indicate that additional verbalized reasoning shifts the prompting baselines slightly toward semantic preservation at the expense of target-style matching.

\begin{table*}[t]
\centering
\begin{tabular}{p{0.3\textwidth}p{0.3\textwidth}p{0.3\textwidth}}
\toprule
\textbf{Input} & \textbf{Target Exemplar} & \textbf{A3S Output} \\
\midrule
Verratti is practically untouchable. He's signing an extension every year or so and PSG won't sell for even a €100m. &
He didn't do any of that ... \textit{oh for fucks sake you are brilliant ... hearing ``DADDDDYYYYY!!''} ... \textit{hehe yeah, ill enjoy it while it lasts ... donuts too? glazed thanks WHERE. WHERE CAN I GET ONE.} &
VerrATTI IS PRACtically UNTACHOOL!!1!! hes SIGNIN A EXTENR ... PSG WONT SELL 4 EVN 100M \\
\addlinespace
You're assuming that Rooney is on the market while it's clear that Mourinho thinks that he can play an important part at this club yet. &
This analogy doesnt really work for jim, per se ... \textit{for chrissakes hes saying BRAVO again! tss tell us how you really feel} ... \textit{lol opies tits sniff} ... &
you're assuming roonee is on the market lol no u r soooo wrong lol \\
\addlinespace
Yeah, and that's the exact way you do a 270 degree Zidane turn (the 4 star turn that u/Ewaninho showed). What Firmino is doing is not a McGeady spin. It's a matter of semantics I guess... And what's up with the ad hominem? &
This analogy doesnt really work for jim, per se ... \textit{for chrissakes hes saying BRAVO again! tss tell us how you really feel} ... \textit{lol opies tits sniff} ... &
yeah and thats the exat way you do a 270 degree zidane turn ... what firmino is doin is not a mcgeeedy spin its a mater of sems i gess and whats up with the ad hominem \\
\bottomrule
\end{tabular}
\caption{Qualitative examples from the MUD dataset. Each case shows the source input, the target exemplar, and the model output. Ellipses indicate truncation for brevity.}
\label{tab:qualitative}
\end{table*}

\section{Human Evaluation}
\label{app:human_eval}

We conduct a pairwise human preference evaluation to complement automatic metrics with direct judgments of (i) stylistic alignment to a given target author exemplar and (ii) preservation of the original meaning. We conduct our evaluation on Qualtrics and use Prolific to source our annotators.

\paragraph{Task overview.}
Each evaluation item consists of two judgments over the same candidate pair:

\begin{itemize}
    \item \textbf{Style transfer success (Part 1).}
    Annotators are shown a target exemplar and two candidate rewrites (Candidate A / Candidate B). They select which candidate best matches the target writing style, with options: \{\textit{Candidate A}, \textit{Candidate B}, \textit{Tie}, \textit{Neither}\}.

    \item \textbf{Rationale elicitation (optional).}
   After the style judgment, annotators optionally indicate why they preferred their choice by multi-selecting one or more stylistic aspects: \textit{Structure}, \textit{Tone}, \textit{Vocabulary}, \textit{Surface (capitalization/formatting)}, \textit{Narrative perspective}, \textit{Figurative language}.

    \item \textbf{Meaning preservation (Part 2).}
    Annotators are shown the original source text and the two candidate rewrites (Candidate A / Candidate B). They are then instructed to select which candidate best preserves the original meaning with the same options: \textit{Candidate A}, \textit{Candidate B}, \textit{Tie}, \textit{Neither}.
\end{itemize}

\paragraph{System labeling.}
Annotators are not informed which system produced which candidate. For each item, we randomize which system's output is displayed as Candidate A and which is displayed as Candidate B.  We record a separate item-level mapping to track which label pertains to which system.

\paragraph{Item sampling and allocation.}
To limit annotation cost, we evaluate on a subset of instances already used for reporting automatic metrics. We sample $N{=}70$ instances using a length-based heuristic (selecting the shortest instances by combined target exemplar, input, and candidate output lengths) to reduce reading burden. Each item receives $J{=}3$ independent judgments, yielding 210 judgments in total. Participants are compensated at an effective rate of approximately \$15/hour, compliant with regional minimum wage.

\begin{figure*}[t]
    \centering
    \includegraphics[width=\textwidth]{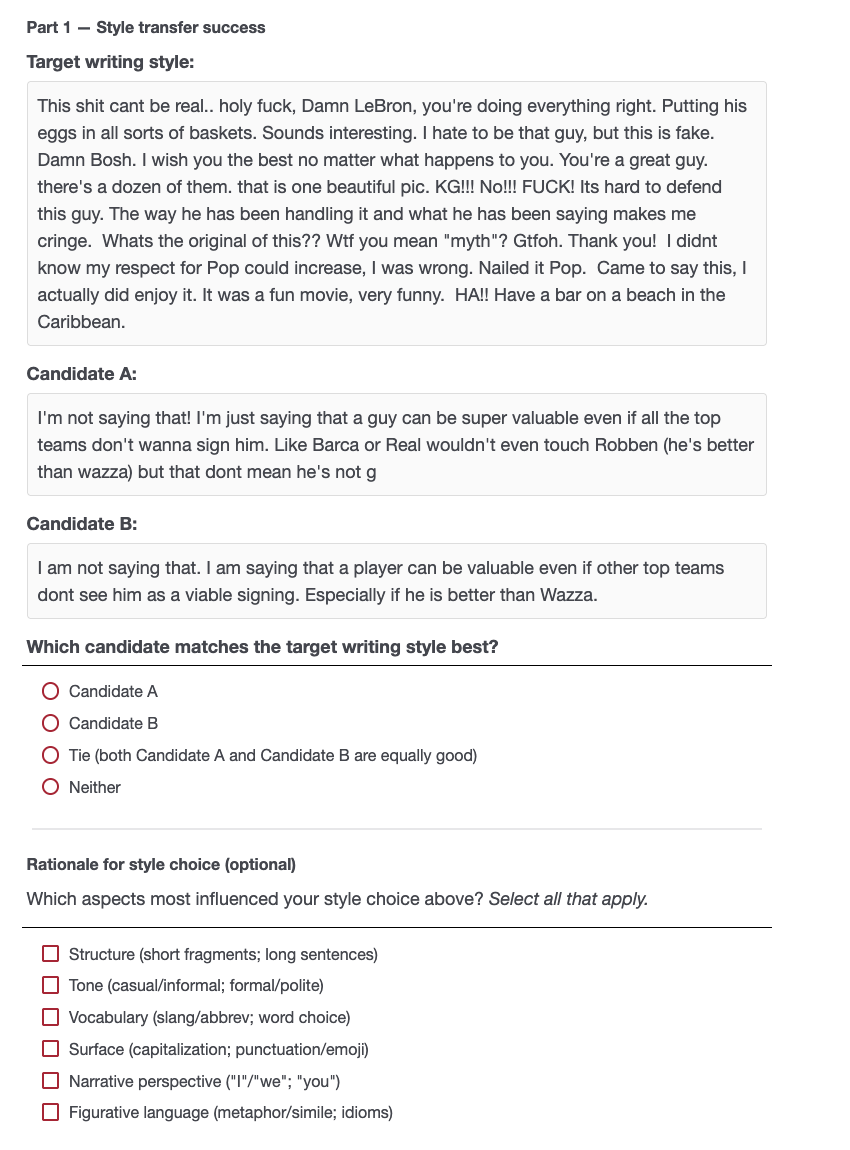}
    \caption{Style transfer success judgment: annotators compare two candidates against the target style exemplar, and then optionally provide a rationale.}
    \label{fig:humaneval_style}
\end{figure*}

\begin{figure*}[t]
    \centering
    \includegraphics[width=\textwidth]{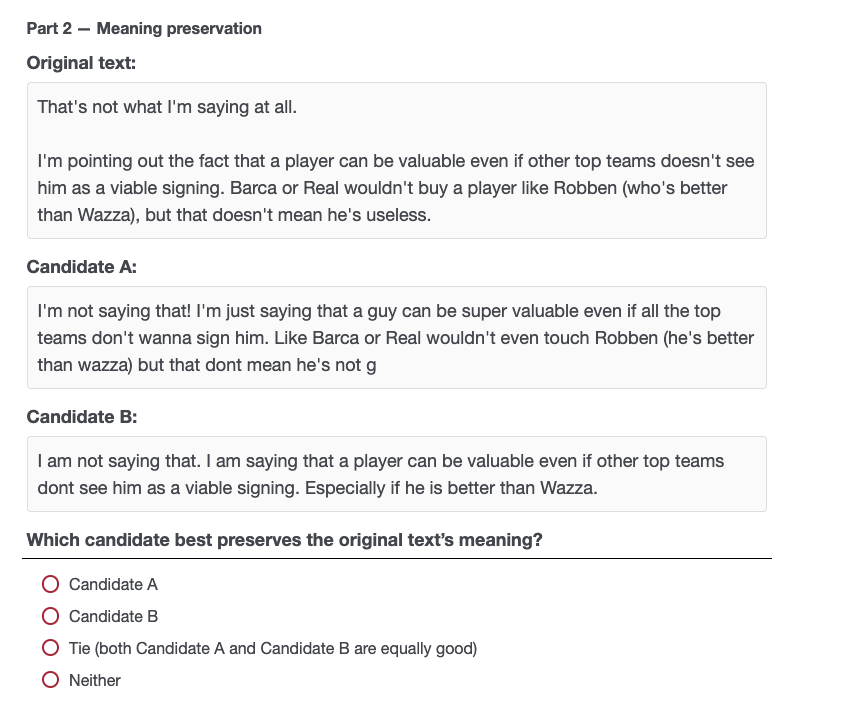}
    \caption{Meaning preservation judgment: annotators compare the same candidates against the original input to assess meaning preservation.}
    \label{fig:humaneval_meaning}
\end{figure*}

\paragraph{Annotator questions.}
Figures~\ref{fig:humaneval_style}--\ref{fig:humaneval_meaning} show screenshots of the survey interface, including the overview instructions, the style judgment (with optional rationale elicitation), and the meaning judgment.

\section{Prompts}

\subsection{LLM-as-a-judge Prompts}
\label{app:llm_prompts}

\begin{figure*}[h]
    \centering
    \small
    \begin{tcolorbox}[
    title={\textbf{Pairwise Style Comparison Prompt}},
    ]
You are an experienced linguist specializing in authorship and writing style analysis.\\
\\
Given a REFERENCE TEXT and two TARGET TEXTS, determine which TARGET TEXT is significantly more likely to have been written by the same author as the REFERENCE TEXT.\\
\\
Base your judgment solely on writing style, including (but not limited to):\\
- Linguistic and stylistic patterns\\
- Sentence structure and organization\\
- Ordering and presentation of information\\
- Tone, formality, and rhetorical habits\\
\\
Do not rely on topic similarity or factual content except where it reflects stylistic choices.\\
\\
For your response, follow these instructions:\\
1. Output TEXT ONE if it is significantly more likely to be written by the same author as the REFERENCE TEXT.\\
2. Output TEXT TWO if it is significantly more likely to be written by the same author as the REFERENCE TEXT.\\
3. Output BOTH if either text could plausibly have been written by the same author, or if neither stands out as more likely.\\
\\
Here are the texts:\\
REFERENCE TEXT:\\
\{reference\_text\}\\
\\
TEXT ONE:\\
\{post\_one\}\\
\\
TEXT TWO:\\
\{post\_two\}\\
\\
First, provide a brief justification for your judgment based on stylistic evidence.\\
Then output exactly one of the following on a new line: TEXT ONE, TEXT TWO, or BOTH.
    \end{tcolorbox}
    \caption{
    The Pairwise Style Comparison prompt used for GPT-4.1 evaluation. This prompt instructs the model to compare two candidate outputs against a reference exemplar solely on stylistic grounds, ignoring topic similarity.
    \label{fig:pairwise_prompt}}
\end{figure*}

The prompt shown in Figure~\ref{fig:pairwise_prompt} determines which of two candidate texts better matches the stylistic signature of the reference. It forces the model to justify its decision based on specific stylistic dimensions before outputting a final verdict.

\subsection{Aspect-Specific Paraphrase Prompts}
\label{app:aspect_prompts}

To compute aspect-specific contrastive activation vectors, we generate paraphrases that isolate individual stylistic dimensions while preserving semantic content. All paraphrases are generated using the same base language model and decoding configuration. Each prompt explicitly instructs the model to preserve meaning and output only the rewritten text.

\begin{figure}[t]
    \centering
    \scriptsize
    \begin{tcolorbox}[
        title={\textbf{Neutral Paraphrase Prompt}},
      width=\columnwidth,
    ]
Rewrite the following passage in a simple, neutral style.
Ensure neutrality in sentence structure, word choice, tone, and overall style.
Avoid figurative language, emotional tone, or persuasive elements.
Output only the rewritten text without explanations or extra content.

Passage:
\{source\_text\}
    \end{tcolorbox}
    \caption{Prompt used to produce neutralized variants.}
    \label{fig:prompt_neutral}
\end{figure}

\begin{figure}[t]
    \centering
    \scriptsize
    \begin{tcolorbox}[
        title={\textbf{Tone Paraphrase Prompt}},
      width=\columnwidth,
    ]
Rewrite the following text in the exact opposite tone. For example, if it's funny, make it serious; if it's angry, make it calm.
Output only the final rewritten text.

Text:
\{source\_text\}
    \end{tcolorbox}
    \caption{Prompt used to generate contrastive tone variants.}
    \label{fig:prompt_tone}
\end{figure}

\begin{figure}[t]
    \centering
    \scriptsize
    \begin{tcolorbox}[
        title={\textbf{Perspective Paraphrase Prompt}},
      width=\columnwidth,
    ]
Rewrite the following text in the opposite narrative perspective/voice.
For example, if it is written in first person (I/me), rewrite it in third person; if it is in third person, rewrite it in first person.
Maintain the original meaning but fully shift the narrative perspective.
Output only the final rewritten text.

Text:
\{source\_text\}
    \end{tcolorbox}
    \caption{Prompt used to generate contrastive narrative perspective variants.}
    \label{fig:prompt_perspective}
\end{figure}

\begin{figure}[t]
    \centering
    \scriptsize
    \begin{tcolorbox}[
        title={\textbf{Vocabulary Paraphrase Prompt}},
      width=\columnwidth,
    ]
Rewrite the following text using a different vocabulary style or register.
Shift the wording to reflect a contrasting linguistic dialect (for example, American vs. British English, simple or complex, formal vs. informal, technical vs. conversational, poetic vs. plain).
Preserve the original meaning.
Output only the rewritten text. No explanations, notes, or additional formatting.

Text:
\{source\_text\}
    \end{tcolorbox}
    \caption{Prompt used to generate contrastive diction/register variants.}
    \label{fig:prompt_vocab}
\end{figure}

\begin{figure}[t]
    \centering
    \scriptsize
    \begin{tcolorbox}[
        title={\textbf{Syntax Paraphrase Prompt}},
      width=\columnwidth,
    ]
Rewrite the following text using the opposite syntax style.
If the text uses long, complex sentences with multiple clauses, embedded phrases, or flowing syntax, rewrite it using short, simple, direct sentences with minimal subordination.
If the text uses short, simple, abrupt, or minimalistic sentences, rewrite it using longer, more complex, syntactically rich sentences with varied structures.
You may rearrange phrases or restructure sentences to achieve the opposite syntax style, but keep the meaning intact.
Output only the rewritten text. No explanations or commentary.

Text:
\{source\_text\}
    \end{tcolorbox}
    \caption{Prompt used to generate contrastive syntactic complexity variants.}
    \label{fig:prompt_syntax}
\end{figure}

\begin{figure}[t]
    \centering
    \scriptsize
    \begin{tcolorbox}[
        title={\textbf{Figurative Language Paraphrase Prompt}},
      width=\columnwidth,
    ]
Rewrite the following text with the opposite figurative and descriptive style.
If the text uses rich imagery, metaphors, similes, or very vivid description, rewrite it in a plain, literal, and minimally descriptive style.
If the text is plain, literal, and low in imagery, rewrite it with frequent figurative language (metaphors, similes, idioms) and rich sensory detail.
Preserve the original meaning.
Output only the rewritten text. No explanations or commentary.

Text:
\{source\_text\}
    \end{tcolorbox}
    \caption{Prompt used to generate contrastive figurative vs.\ literal variants.}
    \label{fig:prompt_figurative}
\end{figure}

\begin{figure}[t]
    \centering
    \scriptsize
    \begin{tcolorbox}[
        title={\textbf{Structure Paraphrase Prompt}},
      width=\columnwidth,
    ]
Rewrite the following text using the opposite structural and formatting style.
Reverse the overall organization, flow, and presentation while preserving the original meaning.

- If the text is long, multi-paragraph, digressive, reflective, or written in a loose or stream-of-consciousness style, rewrite it in a concise, tightly structured, linear, and logically organized format.
- If the text is brief, highly structured, list-based, or presented in a clear linear sequence, rewrite it in a more expansive, exploratory, free-flowing, and digressive style, possibly across multiple paragraphs.

You may change sentence length, paragraph breaks, transitions, and the order of ideas as needed to achieve the opposite structural style, but do not alter the meaning or core content.
Output only the rewritten text. No explanations or extra commentary.

Text:
\{source\_text\}
    \end{tcolorbox}
    \caption{Prompt used to generate contrastive structural/organizational variants.}
    \label{fig:prompt_structure}
\end{figure}

\begin{figure}[t]
    \centering
    \scriptsize
    \begin{tcolorbox}[
        title={\textbf{Surface Form Paraphrase Prompt}},
      width=\columnwidth,
    ]
Rewrite the following text with the opposite surface style and formatting quirks.
If the text uses a lot of emojis, ALL CAPS, unusual spacing, bullet points, or repeated symbols, rewrite it in a clean, standard prose style with normal capitalization and punctuation.
If the text is currently clean, standard prose, rewrite it with distinctive quirks such as emojis, expressive capitalization, unconventional punctuation, and/or bullet lists where appropriate.
Preserve the original meaning, but change only surface form and visible quirks.
Output only the rewritten text. No explanations or commentary.

Text:
\{source\_text\}
    \end{tcolorbox}
    \caption{Prompt used to generate contrastive surface-form variants.}
    \label{fig:prompt_surface}
\end{figure}

\subsection{Aspect-Aware Prompting Baseline}
\label{app:aspect_aware_prompting}

The Aspect-Aware Prompting baseline isolates the rhetorical aspect decomposition used by \AThreeS\ from its activation-space realization. Rather than constructing per-aspect contrastive activation directions, the baseline elicits the same information from the base language model purely through natural-language prompting. Concretely, it runs a three-turn chain (Figures~\ref{fig:prompt_aspect_analysis}--\ref{fig:prompt_aspect_rewrite}) over the same set of rhetorical dimensions used in \AThreeS\ (\textit{Tone, Figurative Language, Vocabulary, Structure, Syntax, Perspective}). The first turn (Figure~\ref{fig:prompt_aspect_analysis}) asks the model to analyze the target exemplar along each aspect. The second turn (Figure~\ref{fig:prompt_aspect_descriptors}) distills that analysis into a comma-separated list of style descriptors that summarize the author's writing style across all six aspects. The third turn (Figure~\ref{fig:prompt_aspect_rewrite}) instructs the model to rewrite the source input to match those descriptors. Conversation history is retained across the three turns. Decoding configuration and the base language model are kept identical to those used to construct \AThreeS's aspect-specific paraphrases (Appendix~\ref{app:aspect_prompts}), so any performance gap between this baseline and \AThreeS\ in Tables~\ref{tab:main_results} and~\ref{tab:main_results_full} reflects the contribution of operating in activation space rather than differences in the underlying aspect inventory, model, or sampling parameters.

\begin{figure}[t]
    \centering
    \scriptsize
    \begin{tcolorbox}[
        title={\textbf{Aspect-Aware Prompting: Turn 1 -- Style Analysis}},
      width=\columnwidth,
    ]
Passage: \{target\_text\}

Analyze the authorship style of this passage along the following aspects:\\
- Tone (e.g., funny, serious, angry, calm, formal, casual)\\
- Figurative language (e.g., use of metaphors, similes, vivid imagery vs. plain literal description)\\
- Vocabulary (e.g., formal vs. informal, technical vs. conversational, simple vs. complex, dialectal register)\\
- Structure (e.g., long meandering vs. concise focused, linear vs. digressive, paragraph and sentence organization)\\
- Syntax (e.g., long complex sentences with multiple clauses and embedded phrases vs. short simple direct sentences with minimal subordination)\\
- Perspective (e.g., first-person, second-person, third-person narrative voice).
    \end{tcolorbox}
    \caption{Turn 1 of the Aspect-Aware Prompting baseline: elicit an aspect-by-aspect analysis of the target exemplar.}
    \label{fig:prompt_aspect_analysis}
\end{figure}

\begin{figure}[t]
    \centering
    \scriptsize
    \begin{tcolorbox}[
        title={\textbf{Aspect-Aware Prompting: Turn 2 -- Style Descriptors}},
      width=\columnwidth,
    ]
Style analysis: \{style\_analysis\}. List some adjectives, comma-separated, that describe the writing style of the author of the target text across the analyzed aspects (tone, figurative language, vocabulary, structure, syntax, perspective). Strictly output only the style descriptors without any other content.
    \end{tcolorbox}
    \caption{Turn 2 of the Aspect-Aware Prompting baseline: distill the per-aspect analysis into a list of style descriptors.}
    \label{fig:prompt_aspect_descriptors}
\end{figure}

\begin{figure}[t]
    \centering
    \scriptsize
    \begin{tcolorbox}[
        title={\textbf{Aspect-Aware Prompting: Turn 3 -- Rewrite}},
      width=\columnwidth,
    ]
Here is a text: \{input\_text\} Rewrite the text to be more \{style\_descriptors\}. Strictly output only the rewritten text without any other content.
    \end{tcolorbox}
    \caption{Turn 3 of the Aspect-Aware Prompting baseline: rewrite the source input to match the elicited style descriptors.}
    \label{fig:prompt_aspect_rewrite}
\end{figure}



\end{document}